%% file: main.tex
\documentclass[10pt]{article}
\usepackage{xnav}

\ifdefined\DRAFT\draftnotestrue\fi

\newcommand{\MissionTakeovers}{}

\reporttitle{X-Nav Technical Report: Real-World Long-Horizon Indoor-to-Outdoor Navigation on a Humanoid Robot}
\reportdate{November 2026}
\reportauthors{\fontsize{9}{12.2}\selectfont
  \au{Jiaxin Huang}{1,2}{\eqc\lead}\quad \au{Chengxing Lin}{1,3}{\eqc}\quad \au{Xiaoyu Luo}{1,4}{\eqc}\quad \au{Sirui Chen}{1,5}{\eqc}\quad \au{Jingyu Guo}{1,6}{\eqc}\quad \au{Aolin Li}{1,7}{\eqc}\\
  \au{Kairui Chen}{1}{}\quad \au{Rui Zhen}{1}{}\quad \au{Cheng Wen}{1}{}\quad \au{Kerui Gu}{1}{}\quad \au{Shunyao Wang}{1}{}\quad \au{Zhihui Peng}{1}{}\quad \au{Shaoli Huang}{1}{\corr}}
\reportaffil{\fontsize{7.8}{9.8}\selectfont
  \textsuperscript{1}AgiBot X-Lab (Boson Lab)\quad \textsuperscript{2}MBZUAI\quad \textsuperscript{3}UESTC\quad \textsuperscript{4}University of Macau \\
  \textsuperscript{5}Nanyang Technological University\quad \textsuperscript{6}The University of Melbourne\quad \textsuperscript{7}Wuhan University}
\reportnotes{\textsuperscript{*}Equal contribution\qquad \textsuperscript{\textdagger}Project leader\qquad \textsuperscript{\corrsign}Corresponding author}

\begin{document}
\makereporttitle
\vspace{-4pt}
\suppressfloats[t]

\input{sections/00_abstract}
\input{sections/01_introduction}
\input{sections/02_overview}
\input{sections/03_data_chengxing}
\input{sections/04_models_chengxing}
\input{sections/05_explicit}
\input{sections/06_workflow}
\input{sections/07_evaluation}
\input{sections/09_related}
\input{sections/10_conclusion}

\bibliographystyle{abbrvnat}
\bibliography{refs}

\appendix
\input{sections/appendix/B_pixelgoal}
\input{sections/appendix/C_runtime}
\input{sections/appendix/D_protocol}
\input{sections/appendix/E_localpolicy}

\end{document}

%% file: sections/00_abstract.tex
%
\begin{firstpagefill}
\vskip 3pt plus 1fill minus 3pt
\begin{abstractbox}
{\sffamily\bfseries\fontsize{8.2}{11}\selectfont\color{aginavy}Abstract.}\;
Long-horizon navigation from indoors to outdoors remains an open problem for humanoid robots: one instruction covers rooms, doors and roads, and the camera moves with the gait.
Worken, AGIBOT's measure of productive robot work, depends on robots that reach their workplaces autonomously.
We present \textbf{\XNav}, to our knowledge the first system to take a bipedal humanoid through a language-instructed long-horizon mission from indoors to outdoors.
\textbf{\XNav} covers a pipeline from data to deployment.
Its \emph{data engine} collects navigation data in open-source scenes, in digital twins of deployment sites and on real robots.
Its navigation model, \textbf{\MEL}, takes only RGB images from the robot's built-in camera and comes in two versions: \textit{\MD} predicts actions, and \textit{\MP} predicts a navigation goal.
Its agentic workflow, \emph{\XAgent}, turns a brief human instruction into sub-tasks and deploys \textbf{\MEL} on the robot with explicit target navigation and obstacle avoidance.
On the \Xtwo, \textbf{\XNav} completes a mission of about a hundred metres from an office lobby to a store in a neighbouring building; LiDAR and odometry are used for target approach and low-level control.
The mission is completed with both versions of \textbf{\MEL}: the robot follows a change of destination and walks through a door from the lobby to the road, and with \textbf{\XAgent} it also passes static and dynamic obstacles on the route.
During deployment, \textbf{\MEL} is adapted with data collected in a digital twin of the site; the adaptation raises its closed-loop success rate in the twin, and the adapted model is then deployed on the robot at the real site.
In addition, \MEL achieves competitive results on the R2R-CE and RxR-CE benchmarks.
\end{abstractbox}
\vskip 12pt plus 1fill minus 8pt
\noindent\begin{minipage}[b]{\linewidth}
\centering
\includegraphics[width=\linewidth]{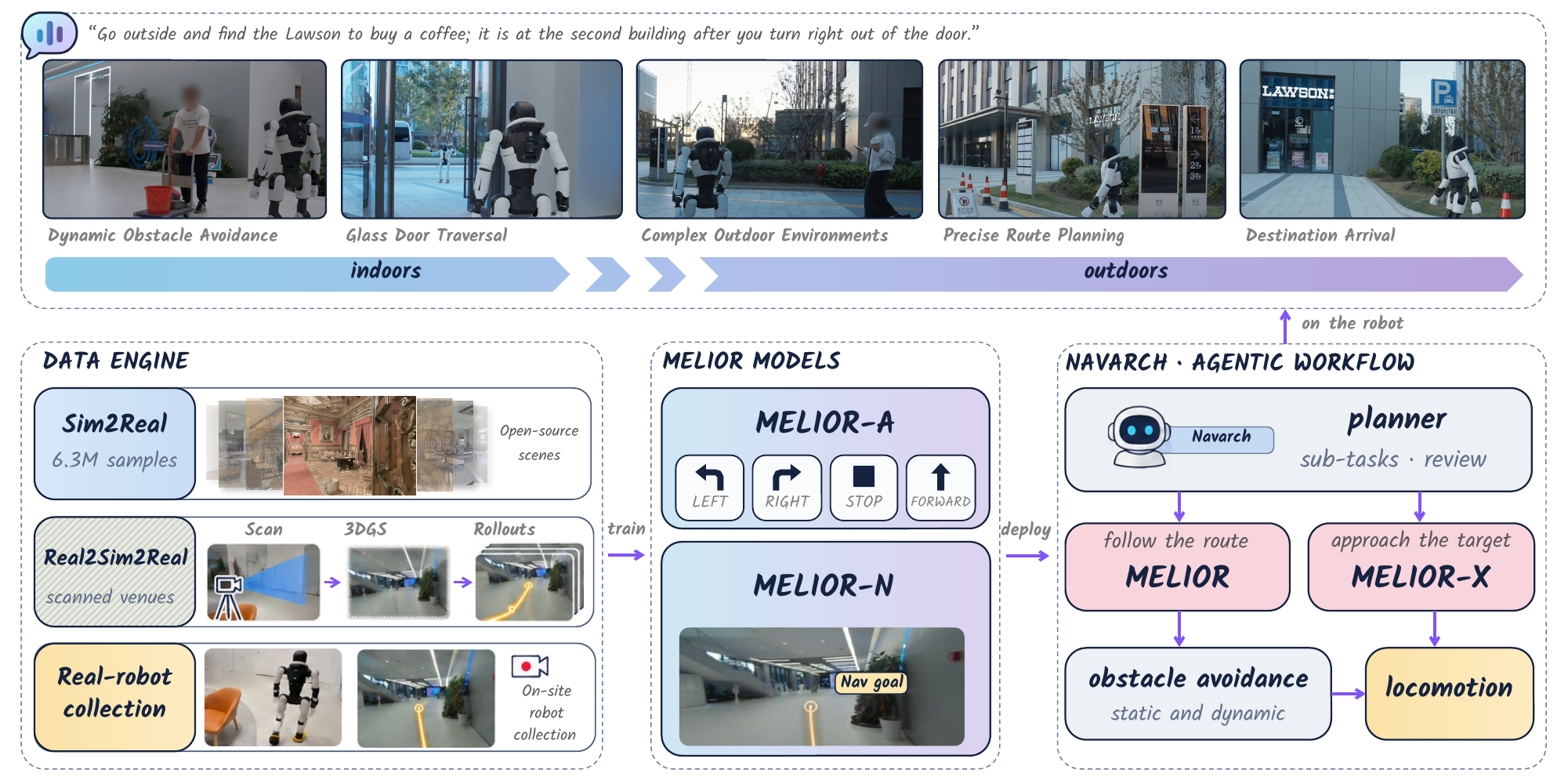}
\captionof{figure}{\textbf{\XNav overview.}
Top: one mission on the \Xtwo, about a hundred metres long, from an office lobby to a store in a neighbouring building, with the instruction that the robot receives; the frames are from the mission with \XAgent and \MP (\cref{fig:agentrun}).
Bottom: the three parts of the system.
The data engine has three collection routes: open-source scene collection (Sim2Real), digital-twin collection at the deployment site (Real2Sim2Real), and real-robot collection.
\MD predicts actions, and \MP predicts a navigation goal (Nav goal).
\XAgent, the agentic workflow, runs \MEL and the explicit navigation skill \MX on the robot with obstacle avoidance.}
\label{fig:teaser}
\end{minipage}
\end{firstpagefill}

%% file: sections/01_introduction.tex
\section{Introduction\label{sec:intro}}

A humanoid robot that works in the real world must first reach its place of work.
Given one instruction, for example to go outside and buy a coffee at a store in a neighbouring building, it has to cross rooms, pass through doors, and walk along roads to its destination.
We refer to such a task as a \emph{mission}.
AGIBOT measures productive robot work by \emph{Worken}, the product of active robots, autonomous working hours, and output efficiency \citep{agibot2026worken}, and each factor depends on robots that reach their workplaces autonomously.
This report studies missions in which a humanoid robot walks about a hundred metres from indoors to outdoors (\cref{fig:teaser}).
Vision-language navigation (VLN) formalises this task: an agent follows a language instruction using its visual observations \citep{anderson2018vln}.
Progress on its benchmarks has been rapid.
Navigation models built on video-based vision--language models have raised the success rate on the continuous-environment benchmarks R2R-CE and RxR-CE year after year \citep{krantz2020vlnce,ku2020rxr,zhang2024navid,zhang2025uninavid,cheng2025navila,wei2026streamvln,zeng2026janusvln}.
On R2R-CE it rose from 37\% to above 70\% within two years (\cref{tab:benchmark}), and on RxR-CE the best reported system now matches the success rate of human followers \citep{zhou2026navgpt3}.
Over the same period, training sets grew to more than ten million samples \citep{zhang2026qwenrobotnav,zhou2026navgpt3}, and real-robot deployments came to include quadrupeds and humanoids \citep{cheng2025navila,wei2026dualvln,chen2026tango}.

\emph{Has embodied navigation had its ``GPT moment''?}
On the benchmarks, it looks that way; on real robots, not yet.
Where a route length is reported, real-robot runs cover a few tens of metres \citep{zhang2026qwenrobotnav,chen2026tango}.
Reported trials on humanoids stay either indoors or outdoors \citep{cheng2025navila,wei2026dualvln,chen2026tango}, and systems that cover larger sites follow a route from map software or from a spatial memory built in advance \citep{tan2026vbotnav,zhou2026holoagent}.
To our knowledge, no reported system has taken a bipedal humanoid through a language-instructed mission of about a hundred metres across the indoor--outdoor boundary.

\emph{Why do high scores in simulation fail to carry over to real robots?}
We trace the gap to three places: the problem that a benchmark poses, the data that models learn from, and the step from a trained model to a walking robot.
We start with the problem: a benchmark episode is a narrower task than a mission.
An episode covers ten to fifteen metres inside a single building.
The scene is a scan, the camera is simulated and stays at a fixed height, every step is executed exactly as commanded, nothing else in the scene moves, and stopping anywhere within three metres of the goal counts as success.
A mission on a humanoid breaks each of these assumptions.
The route is several times as long and leaves the building, where walls a few metres away give way to open ground and buildings tens of metres away.
The camera moves with the gait, and the step that the robot takes can differ from the one commanded.
People, carts, and cars cross its path.
Nor is stopping within three metres good enough: a robot sent to a store has to arrive at its entrance.
A high score therefore shows that a model follows instructions under benchmark conditions; it says little about the conditions that the benchmark leaves out.
We therefore take the mission itself as the problem, and define it as follows.
A bipedal humanoid receives an instruction in natural language, and the instruction may change during the mission.
Its navigation model sees only RGB images from the robot's built-in camera, and the robot has neither a map of the site nor a reference route.
The route is about a hundred metres long, starts indoors, and ends outdoors, with static and dynamic obstacles along the way.
The mission is complete when the robot stops at the destination named in the instruction.

\emph{Will more data close the gap?}
Scaling up the training set has been the usual response \citep{zhang2026qwenrobotnav,zhou2026navgpt3}, and it does push benchmark scores up.
But the data are the second place where the gap opens, because a sample can teach only what it contains.
Standard VLN data are collected in scanned indoor scenes with a simulated camera \citep{chang2017matterport3d}, the same kind of setting that the benchmarks use.
Such samples contain neither the site where the robot will work nor the robot's own camera, gait, and stopping behaviour, and adding more of them cannot supply what is missing.
Reference routes also show only correct behaviour, whereas a model on a long route has to recover from its own mistakes.
More is better only when it adds what is missing: the first question is what kind of data, not how much.

\emph{What kind of data does a humanoid need, and where should it come from?}
Working back from what the model has to learn, we find three needs, each met by a different source.
First, the model has to follow instructions across many kinds of scenes and routes.
This calls for scale and diversity, and open-source scenes in simulation supply both.
Second, the model has to work at the site where the robot is deployed.
This calls for the layout and appearance of that site: a scan turns the site into a digital twin, in which routes are collected and the model is run so that its mistakes can be corrected \citep{ross2011dagger} before the robot sets foot there.
Third, the model has to work on this particular robot.
This calls for the robot's own camera views and motion, which recordings on the real robot capture directly.

In simulation and in the twin, each sample pairs the camera view with an instruction and with supervision that matches both: the next actions or a navigation goal.
The three sources differ in size by orders of magnitude, from 6.3M samples in open-source scenes to a few hundred routes at a single site, and each supplies what the others lack.
As for how much is enough, we judge it where the robot works, in closed loop in the twin and on the robot, rather than on the benchmark alone.
In the twin of an office lobby, a few hundred routes adapt the model to the site, and adding correction data raises its closed-loop success rate from 40\% to 80\% (\cref{sec:eval_r2s2r}).

\emph{What does it take to put the model on a real robot?}
Deployment is the third place where the gap opens: a trained model is not yet a robot that walks.
The first decision is the camera.
Real-robot deployments commonly mount an additional camera on the robot, such as an Intel RealSense D455 \citep{zhang2025uninavid,wei2026streamvln,wei2026dualvln}; a model that uses the robot's built-in camera leaves the hardware untouched.
The rest is what a mission asks for beyond the output of a model: a brief instruction has to be divided into sub-tasks, each stop has to be checked, a predicted goal has to stay valid while the robot keeps walking, static and dynamic obstacles have to be avoided, and the last few metres to a named target have to be covered.

\emph{What, in the end, should a framework for real-world humanoid navigation look like?}
It has to close the gap in all three places: to take the mission, rather than the benchmark episode, as its problem; to collect data that contain the site and the robot; and to carry the model all the way onto the robot.
Such a framework is more than a model: it covers the full pipeline from data to deployment, and it is tested on the robot itself.

This report presents \textbf{\XNav}, a navigation system built along these lines (\cref{fig:teaser}).
The \emph{data engine} collects navigation data from the three sources: open-source scene collection at scale, digital-twin collection at the deployment site, and real-robot collection on the \Xtwo humanoid.
\emph{\MEL}, the navigation model, takes only RGB images from the built-in camera of the \Xtwo and requires no additional camera.
It comes in two versions: \MD predicts actions, and \MP predicts a navigation goal.
\emph{\XAgent}, the agentic workflow, deploys \MEL on the robot: a planner divides a brief human instruction into sub-tasks and reviews each stop, \MEL follows each route, the explicit navigation skill \MX approaches a named target over the last metres, and an obstacle-avoidance layer operates between the navigation goal and the locomotion controller.
\XAgent uses no map of the site; LiDAR and odometry are used for target approach and low-level control.
On the \Xtwo, \XNav completes a mission of about a hundred metres from an office lobby to a store in a neighbouring building with both versions of \MEL: a run of 96.6\,m with \MD, and a run with \XAgent and \MP that encounters static and dynamic obstacles (\cref{sec:eval_mission}).
In an office lobby, the data engine and \MD close the loop from a scan to the robot: the model is adapted with data collected in the digital twin of the lobby, and the adapted model drives the \Xtwo in the real lobby (\cref{sec:eval_r2s2r}).
On R2R-CE and RxR-CE, \MEL is competitive with recent models that use one forward camera (\cref{sec:eval_benchmark}).
Our contributions are as follows.

\begin{itemize}
\item We present \textbf{\XNav}, to our knowledge the first navigation system that takes a bipedal humanoid through a language-instructed long-horizon mission from indoors to outdoors.
\item We build a data engine that collects navigation data in open-source scenes, in digital twins of deployment sites, and on real robots, and that supports the adaptation of the model to a deployment site.
\item We develop \textbf{\MEL}, a navigation model that takes only RGB images from the robot's built-in camera and predicts actions (\MD) or a navigation goal (\MP), and \textbf{\XAgent}, an agentic workflow that deploys it on the \Xtwo with explicit target navigation and obstacle avoidance.
\end{itemize}

\Cref{sec:overview} gives an overview of the system.
\Cref{sec:data} describes the data engine, \cref{sec:models} the two versions of \MEL, \cref{sec:explicit} the explicit navigation skill, and \cref{sec:workflow} \XAgent.
\Cref{sec:evaluation} reports the experiments, \cref{sec:related} reviews related work, and \cref{sec:conclusion} concludes.

%% file: sections/02_overview.tex
\section{\XNav Overview\label{sec:overview}}

\providecommand{\lbl}[2]{\tcbox[on line,colback=#1,colframe=#1,boxrule=0pt,arc=3pt,left=3pt,right=3pt,top=0.5pt,bottom=0.5pt]{\sffamily\fontsize{6.6}{8}\selectfont\bfseries #2}}
\providecommand{\Lone}{\lbl{policyworld}{L1}\xspace}
\providecommand{\Ltwo}{\lbl{frozenworld}{L2}\xspace}
\providecommand{\Lthree}{\lbl{dataworld}{L3}\xspace}

\XNav consists of three parts (\cref{fig:teaser}).
The \emph{data engine} collects navigation data through three complementary routes: open-source scene collection, digital-twin collection and real-robot collection.
\emph{\MEL}, the navigation model, takes an instruction and RGB frames from the built-in camera of the \Xtwo and predicts actions (\MD) or a navigation goal (\MP).
\emph{\XAgent}, the agentic workflow, divides a mission into sub-tasks, runs the navigation skills on the robot, and passes their goals through obstacle avoidance to the locomotion controller.

Real-robot deployments of navigation models commonly mount an additional camera on the robot, for example an Intel RealSense D455 \citep{zhang2025uninavid,wei2026streamvln,wei2026dualvln}.
\XNav instead uses the cameras that the \Xtwo already has: \MEL navigates with the built-in head-mounted RGB camera, \MX additionally uses the built-in RGB-D camera, and no camera is added to the robot.

\Cref{tab:landscape} compares \XNav with recent navigation systems.
These systems predict waypoints or actions from RGB input, and those that cover larger sites follow a route or a spatial memory that is prepared in advance.
The navigation model of \XNav takes a single RGB stream, the system uses no site prior, and it is demonstrated on a mission of about a hundred metres from indoors to outdoors on a bipedal humanoid.

\begin{table}[!ht]
\centering\small
\caption{\textbf{\XNav and recent navigation systems.}}
\label{tab:landscape}
\setlength{\tabcolsep}{5pt}
\begin{tabularx}{\linewidth}{@{}l >{\raggedright\arraybackslash}p{2.7cm} l l >{\raggedright\arraybackslash}X@{}}
\toprule
\tablehead{System} & \tablehead{Output} & \tablehead{Input} & \tablehead{Site prior} & \tablehead{Highlight}\\
\midrule
Qwen-RobotNav \citeyearpar{zhang2026qwenrobotnav} & waypoints & RGB & none & 15.6M training samples\\
Robostral Navigate \citeyearpar{majumdar2026robostral} & image waypoints & single RGB & none & reinforcement learning\\
NavGPT-3 \citeyearpar{zhou2026navgpt3} & tool calls, actions & RGB, map & none & reasoning planner above an action policy\\
VbotNav-0 \citeyearpar{tan2026vbotnav} & sub-goal, trajectory & RGB, LiDAR & route or memory & cloud--edge policy, indoor and outdoor\\
HoloAgent-0 \citeyearpar{zhou2026holoagent} & skill calls & RGB or LiDAR & spatial memory & persistent 3D spatial memory\\
\oursrow \XNav (ours) & actions or navigation goal & single RGB & none & $\sim$100\,m indoor-to-outdoor humanoid mission\\
\bottomrule
\end{tabularx}
\end{table}

The rest of this section describes the three loops in which the parts work together (\cref{sec:loops}), lists the components (\cref{sec:names}), distinguishes the two types of sub-task in a mission (\cref{sec:modes}), and describes the robot (\cref{sec:platform}).

\subsection{Three loops\label{sec:loops}}

\begin{figure}[!ht]
\centering
\includegraphics[width=\linewidth]{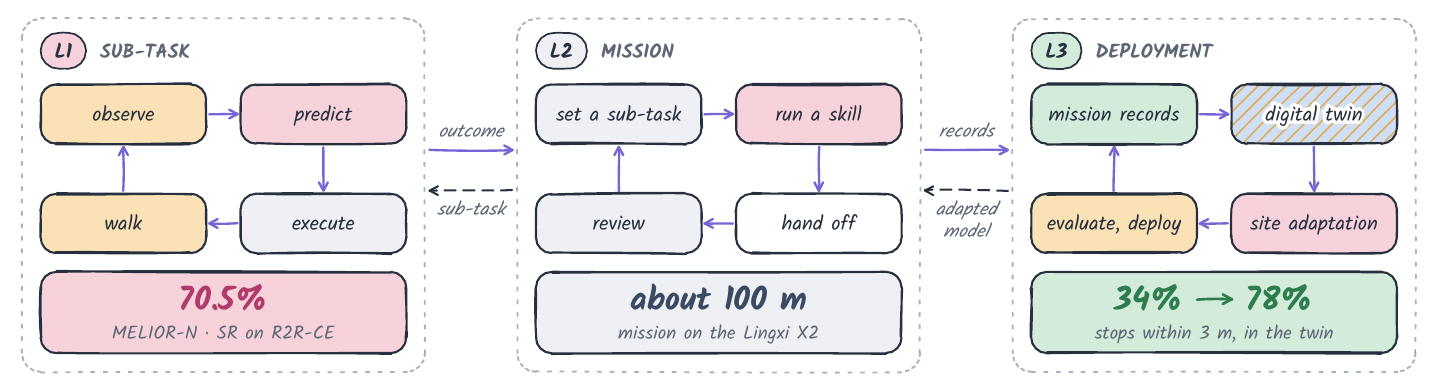}
\caption{\textbf{The three loops of \XNav.}
The sub-task loop (L1), the mission loop (L2) and the deployment loop (L3).
The box below each loop gives one result of this report: the success rate (SR) of \MP on R2R-CE, the length of the mission on the \Xtwo, and the share of closed-loop episodes in a digital twin that stop within 3\,m of the goal without and with correction windows.}
\label{fig:loops}
\end{figure}

The three parts work together in three nested loops (\cref{fig:loops}).

\begin{description}[leftmargin=1.4em,style=sameline,itemsep=3pt]
\item[\Lone Sub-task.]
\MEL observes RGB frames from the built-in camera, predicts actions or a navigation goal, and the robot executes the prediction and walks.
The loop repeats until the model predicts a stop or the sub-task reaches a limit (\cref{sec:workflow_mission}).
\item[\Ltwo Mission.]
The planner sets a sub-task and delegates it to a navigation skill, \MEL for a route and \MX for a named target.
At hand-off the planner reviews the outcome and continues with the next sub-task, corrects the current one, or ends the mission (\cref{sec:workflow}).
\item[\Lthree Deployment.]
The deployment site is reconstructed as a digital twin, and the model is adapted to the site with reference trajectories and correction windows collected in the twin.
The adapted model is evaluated in closed loop in the twin before it is deployed.
\XAgent records every mission, and the records can guide subsequent collection by identifying failed episodes to replay and new areas to scan (\cref{sec:r2s2r}).
\end{description}

\subsection{Components and names\label{sec:names}}

\Cref{tab:names} lists the components.
\XNav is the name of the system, and \MEL the name of its navigation model: \MD is the version that predicts actions, and \MP the version that predicts a navigation goal.
The two versions share the input contract and differ in the output (\cref{sec:contract}).

\begin{table}[!ht]
\centering\small
\caption{\textbf{Components of \XNav.}}
\label{tab:names}
\begin{tabularx}{\linewidth}{@{}l l >{\raggedright\arraybackslash}X@{}}
\toprule
\tablehead{Name} & \tablehead{Part} & \tablehead{Description}\\
\midrule
\XSim & data engine & Simulation platform for data collection and evaluation.\\
\XSyn & data engine & Navigation data collected in open-source scenes.\\
\XTwin & data engine & Navigation data collected in the digital twin of a deployment site.\\
\XReal & data engine & Onboard RGB and robot motion recorded on the \Xtwo.\\
\midrule
\MD & navigation model & Predicts up to four actions per decision.\\
\MP & navigation model & Predicts one navigation goal per decision.\\
\midrule
Planner & \XAgent & Divides a mission into sub-tasks and reviews their outcome.\\
\MX & \XAgent & Explicit navigation skill that approaches a named target.\\
Goal adapter & \XAgent & Converts a navigation goal into a goal pose for the robot.\\
Obstacle-avoidance layer & \XAgent & Avoids static and dynamic obstacles on the way to the goal pose.\\
\bottomrule
\end{tabularx}
\end{table}

\subsection{Following a route and approaching a target\label{sec:modes}}

A mission contains two types of sub-task.
The first is a route, for example ``cross the lobby, leave through the glass door and turn right''.
The second is a target, for example a vending machine, a blue sofa or a store entrance.
\XNav handles the two types with different navigation skills (\cref{tab:modes} in \cref{app:runtime}). \MEL follows a route by \emph{implicit navigation}.
A route description is long and compositional, its landmarks are open-vocabulary, and the correct action depends on the whole view.
The model therefore maps the instruction and the recent frames directly to actions or to a navigation goal, and the route is represented implicitly, in the weights and the context of the model. \MX approaches a target by \emph{explicit navigation}.
An end-to-end model reaches the vicinity of a destination more often than it stops there (the gap between OSR and SR in \cref{tab:benchmark}), and its representation of the target cannot be inspected when the object is small, partly occluded or observed from a changing viewpoint.
The skill therefore maintains an explicit target state, a mask $M_t$ in the image and a position $\mathbf{p}^B_t$ in the robot base frame, and computes the motion from this state.
In a mission, route following covers tens of metres and the approach covers the last few metres.

\subsection{Robot and compute\label{sec:platform}}

All real-robot results use the \Xtwo, a bipedal humanoid with two built-in cameras, a head-mounted RGB camera and a downward-pitched RGB-D camera, a LiDAR and leg odometry.
\MEL takes only the RGB stream of the head-mounted camera, rectified to $640\times480$ with a horizontal field of view of $79^\circ$.
The RGB-D camera, the LiDAR and odometry are used for target approach and low-level control, not by \MEL.
\MX uses the head-mounted camera as its standard view, estimates the position of the target from LiDAR points, and switches to the RGB-D camera close to the target (\cref{sec:explicit}).
The goal adapter uses LiDAR and odometry to obtain the range of a predicted pixel goal and to keep the goal fixed while the robot moves (\cref{sec:goal_adapter}).
An obstacle-avoidance layer operates between the goal pose and the locomotion controller (\cref{sec:avoidance}); walking and turning are executed by the locomotion controller of the \Xtwo.
Inference runs on an off-board GPU that exchanges frames and goals with the robot over the network.

%% file: sections/03_data_chengxing.tex
\section{Data Engine: From Synthetic Scale to the Real Robot}\label{sec:data}

\begin{figure}[!t]
\centering
\includegraphics[width=\linewidth]{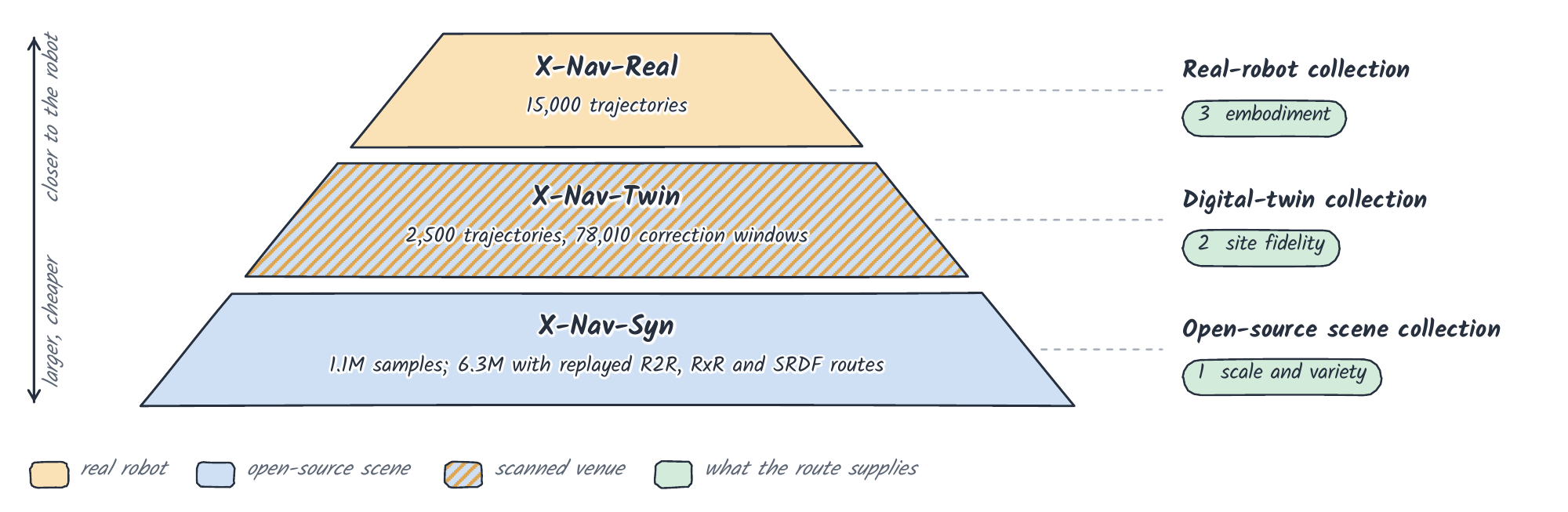}
\caption{\textbf{The three routes to navigation data.}
Collection in open-source scenes covers diverse layouts and tasks.
Collection in the digital twin supplies data for site adaptation.
Collection on the \Xtwo{} captures the robot's onboard observations
and physical motion.
Each tier is labelled with its dataset, \XSyn{}, \XTwin{}, or \XReal{}
(\cref{sec:tiers}); the base tier gives the samples of \XSyn{} and the total of
open-source scene collection, which includes the replayed routes of R2R, RxR,
and SRDF.}
\label{fig:datapyramid}
\end{figure}

A humanoid navigation policy needs data that cover diverse scenes and
tasks, reflect the appearance and layout of the deployment site, and
capture the physical robot's camera observations, gait, and stopping
behaviour. We collect these data for \XNav{} through three complementary
routes (\cref{fig:datapyramid}).

\begin{itemize}
    \item \textbf{Open-source scene collection.}
    We generate trajectories at scale to cover diverse layouts, targets,
    and navigation tasks.

    \item \textbf{Digital-twin collection.}
    We reconstruct a deployment site from a scan and collect trajectories
    and correction windows in its digital twin.

    \item \textbf{Real-robot collection.}
    We record onboard observations and executed motion directly on the
    physical \Xtwo{}.
\end{itemize}

\subsection{\XSim{} platform design}

For the first two routes, we need to replay planned trajectories for
data collection and evaluate learned policies in the same scenes. We
designed \XSim{}, a navigation simulation platform built on Isaac Sim \citep{NVIDIA_Isaac_Sim},
to support both workflows. It provides common interfaces for
observations, policy inference, execution of actions or navigation goals,
state logging, and metric computation.

\paragraph{Execution configurations}
We use a pose-controlled camera for batch trajectory replay and policy
benchmarking. For collection and validation with the simulated \Xtwo{},
we plan routes using the robot's footprint and execute commands through
its locomotion controller.

\subsection{Trajectory collection and annotation}\label{sec:collection}

\begin{figure}[!t]
\centering
\includegraphics[width=\linewidth]{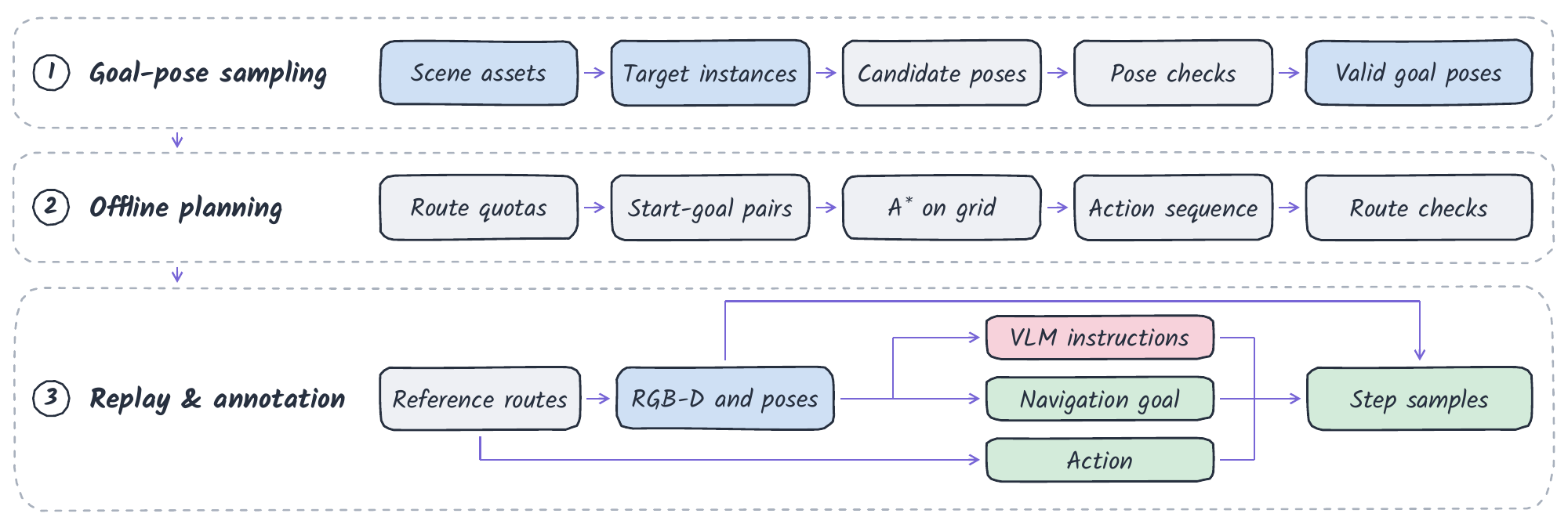}
\caption{\textbf{The generation pipeline of \XSyn.}
(1) \emph{Goal-pose sampling.} We select target instances and sample
goal poses that pass reachability, visibility, and clearance checks.
(2) \emph{Offline planning.} We sample start-goal pairs within route
quotas, plan paths with A* on the occupancy grid, convert them to action
sequences, and check the resulting routes.
(3) \emph{Replay \& annotation.} We replay the reference routes to record
RGB-D observations and camera poses. A vision--language model generates
instructions. We derive action labels from the routes and navigation
goal labels from the replayed poses and depth. We combine the
observations, instructions, and labels into step samples.}
\label{fig:pipeline}
\end{figure}

\begin{figure}[!t]
\centering
\includegraphics[width=\linewidth]{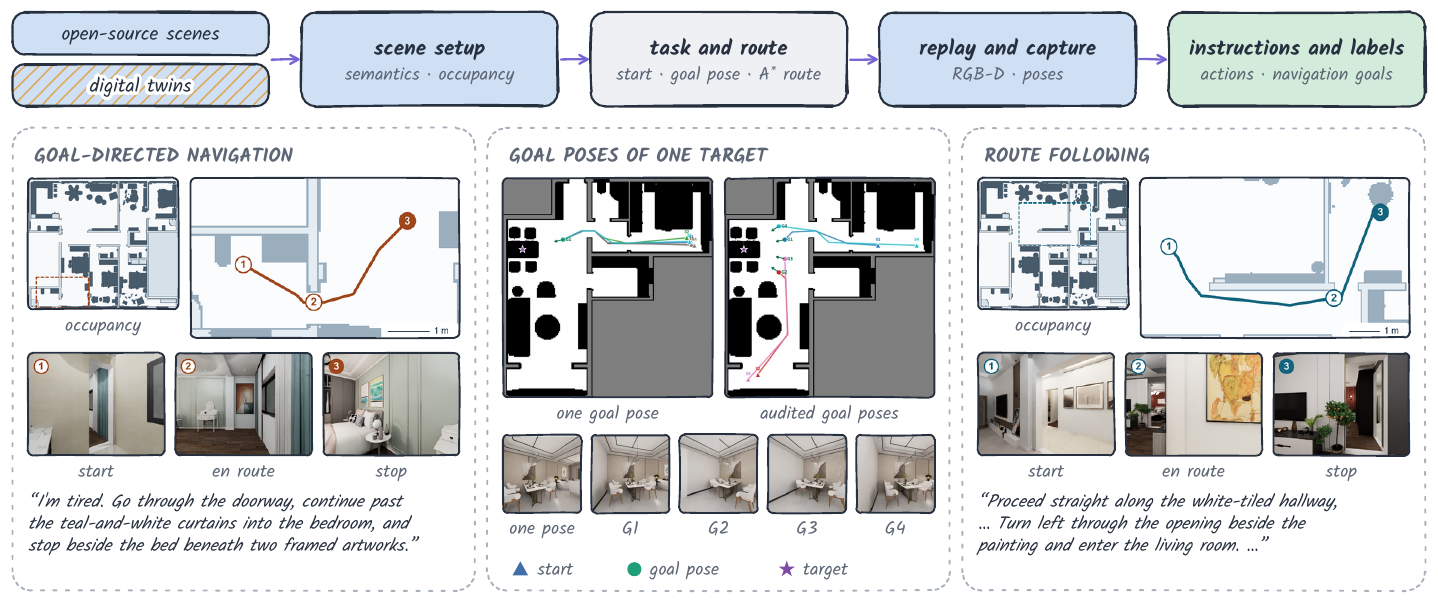}
\caption{\textbf{Trajectory collection in \XSim{}.}
Top: the collection steps, shared by open-source scenes and digital twins
(\cref{fig:pipeline} gives the details).
Left and right: one collected example of each task type, goal-directed
navigation and route following, with the occupancy grid of the scene, the
reference route in the region of the dashed box, the first-person views at the
three numbered positions, and the generated instruction.
Middle: goal poses of one target, a dining table. With one goal pose, routes
from different starts end at the same pose and share the final view. With the
goal poses that pass the pose checks, labelled audited goal poses (G1--G4),
routes from different directions end at different poses, each with the target
in view.
Triangles mark starts, circles indicate goal poses and headings, and
the star marks the target.}
\label{fig:goalposes}
\end{figure}

Each training example needs observations, a route instruction, and
supervision that agree with one another. We plan reference routes
offline and replay them in \XSim{} to collect and label the examples
along the same trajectory (\cref{fig:pipeline}).
\Cref{fig:goalposes} summarises these steps for both scene sources,
open-source scenes and digital twins, and shows collected examples.

\paragraph{Goal-pose sampling}

Routes to the same object can share the same final view even when they
start from different places: with a single goal pose per target, they all
end at that pose (\cref{fig:goalposes}, middle, ``one goal pose''). To cover
more approaches, we select a target instance from the scene annotations and
check several candidate goal poses around it. We check whether each pose is reachable, has
enough room to stop, and lets the robot face and see the target. The
checks account for distance, viewing angle, line of sight, occlusion,
and truncation. We sample the poses that pass in rotation when planning
routes to the same target, so that routes from different directions end at
different goal poses, each with the target in view (G1--G4 in
\cref{fig:goalposes}, middle).

\paragraph{Offline planning}

Repeated sampling in the same scenes and around the same targets can
produce similar routes and views. To broaden coverage, we allocate
routes across scenes, task types, target categories, and length ranges.
The task types are goal-directed navigation, in which the instruction
leads to a target object that it names or implies through a stated need, and
route following, in which the instruction describes the route step by step
(\cref{fig:goalposes}, left and right).
We build an occupancy grid from the scene geometry and plan routes
with A* using a clearance preference of 0.35\,m. We smooth each path
and convert it into 0.25\,m forward steps and $15^\circ$ turns.
Before accepting a route, we check its swept path for collisions and
verify its final pose and stop label.

\paragraph{Trajectory replay and instruction annotation}

We replay the planned pose sequence to render first-person RGB-D views
from the scene assets and record the corresponding camera poses. A
vision--language model generates a navigation instruction from
8--24 keyframes, the route map, and target metadata.
Each example in \cref{fig:goalposes} shows the reference route on the
occupancy grid, three of the rendered views, and the generated instruction.

\paragraph{Supervision and export}

At each decision step, we derive two forms of supervision from the
reference trajectory, one for each version of \MEL{} (\cref{sec:models}).

\begin{itemize}
    \item \textbf{Action.} We use the next action or a short sequence
    of actions along the reference route.

    \item \textbf{Navigation goal.} We select a future pose on the
    reference route and record its position and heading relative to
    the agent's current pose. We also record the goal's image coordinates
    when its projection falls within the current view and passes a
    depth-based occlusion check. A label can also specify a turn in
    place or a stop.
\end{itemize}

We pair each label with its observations and the route instruction,
then check that episode records are complete before merging the
exported samples.

\subsection{Collected navigation data}\label{sec:tiers}

\subsubsection{Open-source scene collection}\label{sec:opensource}

To cover a range of indoor layouts and goal objects, we build \XSyn{}
by sampling new tasks, planning reference routes, and replaying them
in \XSim{}. The \XSyn{} subset used for training provides
1,113,803 step-level samples from 23,761 trajectories across 1,095 scenes.
It includes 461,027 samples from 176 synthetic indoor scenes in VLNVerse \citep{lin2025vlnverse}
and 652,776 samples from 919 reconstructed SAGE-3D \citep{miao2026towards} scenes.
We also replay existing reference routes in \XSim{} to obtain 636,949
samples from R2R \citep{krantz2020vlnce}, 1,889,162 from RxR \citep{ku2020rxr}, and 2,670,750 from SRDF \citep{wang2025bootstrapping}. Together
with \XSyn{}, these sources provide 6,310,664 step-level navigation
samples for supervised fine-tuning (\cref{sec:training_data}).

\begin{figure}[t]
\centering
\includegraphics[width=\linewidth]{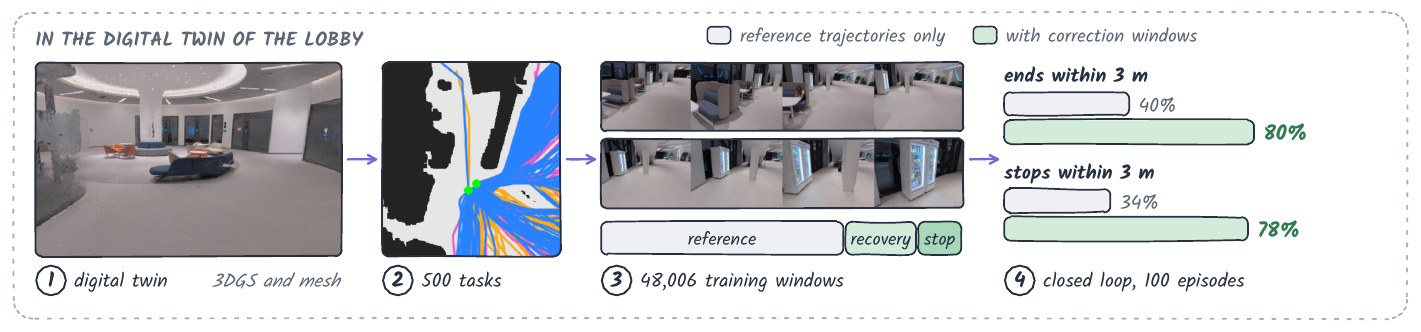}
\caption{\textbf{Digital-twin collection in an office lobby.}
(1) The digital twin of the scanned lobby, rendered with 3D Gaussian splatting.
(2) 500 tasks planned on the occupancy grid of the twin; the lines are
reference trajectories.
(3) The 48,006 training windows of the 500-trajectory setting: windows from
reference trajectories, recovery windows, and stop-boundary windows.
(4) Closed-loop evaluation in the twin for an adapter trained on reference
trajectories only and for one trained with correction windows
(\cref{sec:eval_r2s2r}).}
\label{fig:s2r2s}
\end{figure}

\subsubsection{Digital-twin collection}\label{sec:r2s2r}

Adapting a policy to a specific site requires examples that reflect its
layout, visual appearance, and named targets. A digital twin lets us
collect reference trajectories under these conditions and label recovery
and stopping errors from the current policy's rollouts; we refer to the
data collected in a twin as \XTwin{}.

We scan the deployment site with a handheld device that records RGB
images, a point cloud, and synchronised poses. We import the reconstructed
site into \XSim{}, using 3D Gaussian splatting for visual appearance
\citep{kerbl2023gaussian} and a mesh with an occupancy layer for depth and
traversability. A manually calibrated object graph in the twin of the office lobby
identifies 35 named objects. We place goal poses in front of the
targets with sufficient clearance for the robot to stop.
\Cref{fig:s2r2s} shows the twin of the office lobby, the tasks planned
in it, and the resulting training windows.

We first collected 500 trajectories to vending machines in the office
lobby, with a 400/50/50 train, validation, and test split. For site adaptation, we used 2,500 reference trajectories. The 500-trajectory setting uses 48,006 resampled training windows: 32,404 from reference trajectories, 9,601 recovery, and 6,001 stop-boundary windows. The recovery and stop-boundary windows are the correction windows, 15,602 in this setting. The 2,500-trajectory setting uses five times these counts: 240,030 windows in total, comprising 162,020 from reference trajectories, 48,005 recovery, and 30,005 stop-boundary windows, that is, 78,010 correction windows. The mixture proportions are 67.5\%, 20.0\%, and 12.5\%, respectively. We obtained the correction windows by rolling out the current policy
in the twin and using a reference planner to label recovery from route
deviations and correct stopping decisions \citep{ross2011dagger}.
\Cref{sec:eval_r2s2r} reports the closed-loop evaluation in the twin for
adapters trained in the 500-trajectory setting.
Mission records can also guide subsequent collection by identifying
failed episodes to replay and new areas to scan.

\begin{figure}[t]
\centering
\includegraphics[width=\linewidth]{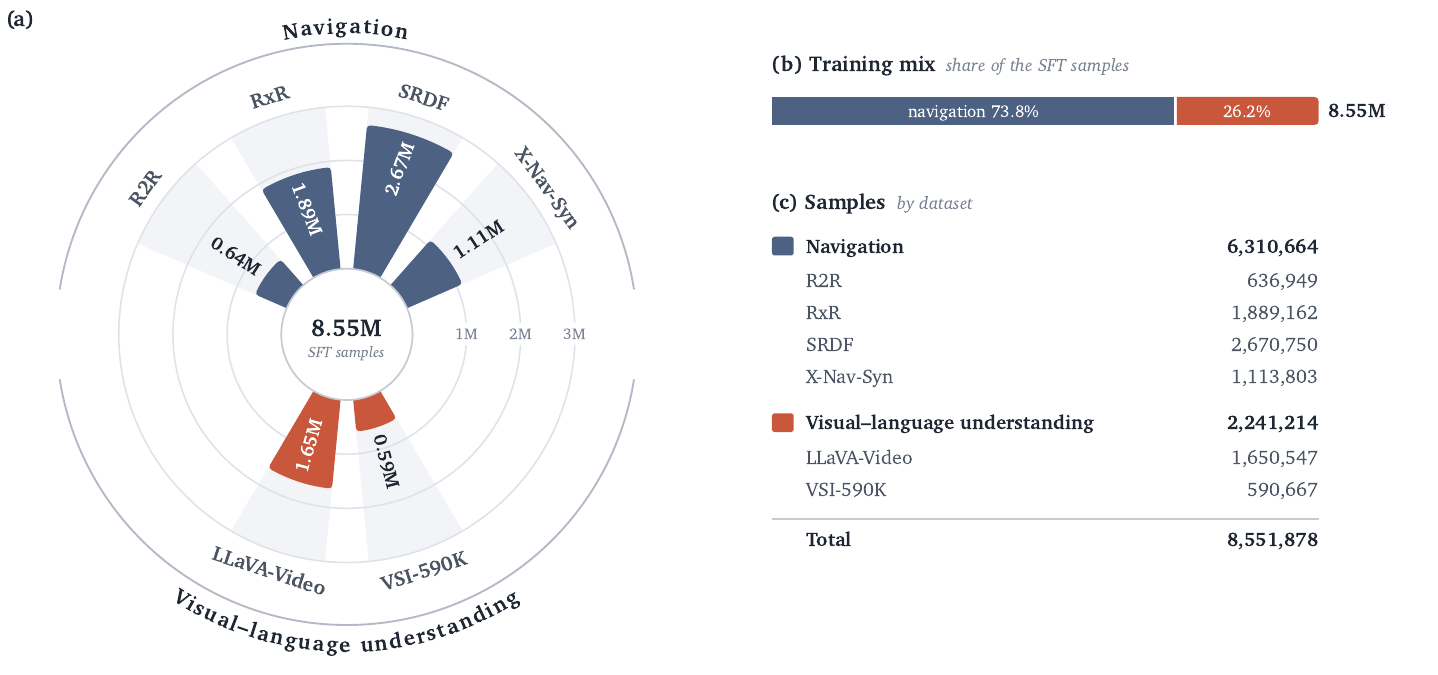}
\caption{\textbf{Supervised fine-tuning data for \MP{}.}
(a) Samples per dataset on a radial axis (rings at 1M, 2M, and 3M
samples), with the four navigation datasets at the top, the two
visual--language understanding datasets at the bottom, and the total in the
centre.
(b) Training mix: navigation samples account for 73.8\% and visual--language
understanding samples for 26.2\% of the 8.55M samples.
(c) Sample counts by dataset.}
\label{fig:training_data}
\end{figure}

\subsubsection{Real-robot collection}\label{sec:realrobot}

On the physical robot, the gait moves the head-mounted camera and
an executed step can differ from the commanded motion. To capture these
effects and the actual stopping position, we collect \XReal{} directly
on the \Xtwo{} with in-house software that records onboard RGB
observations and robot motion. The software supports concurrent
collection with multiple robots at the same site. The target
categories include trash cans, coffee machines, and vending machines.

\subsection{Training data}\label{sec:training_data}

The supervised fine-tuning (SFT) dataset for \MP{} contains 8,551,878
samples (\cref{fig:training_data}). The 6,310,664 navigation samples
described in \cref{sec:opensource} (73.8\% of the total) provide
step-level supervision for predicting navigation goals. We mix in
1,650,547 video understanding examples from LLaVA-Video
\citep{zhang2025llavavideo} and 590,667 spatial reasoning examples from
VSI-590K \citep{yang2025cambrians}, together 26.2\% of the total, to help retain the
backbone's visual--language capabilities during navigation fine-tuning.
\Cref{fig:training_data} gives the samples per dataset (a), the shares of
the two groups (b), and the exact counts (c); SRDF, RxR, and LLaVA-Video are
the three largest sources.

%% file: sections/04_models_chengxing.tex
\section{\MEL: The Navigation Models\label{sec:models}}

\MEL is the navigation model of \XNav.
It follows a route that is given in language, and it comes in two versions that differ in their output: \MD predicts actions, and \MP predicts a navigation goal.
This section gives the input contract the two share, the two output interfaces, and the training.

\subsection{One input contract, two interfaces\label{sec:contract}}

\begin{figure}[t]
\centering
\includegraphics[width=\linewidth]{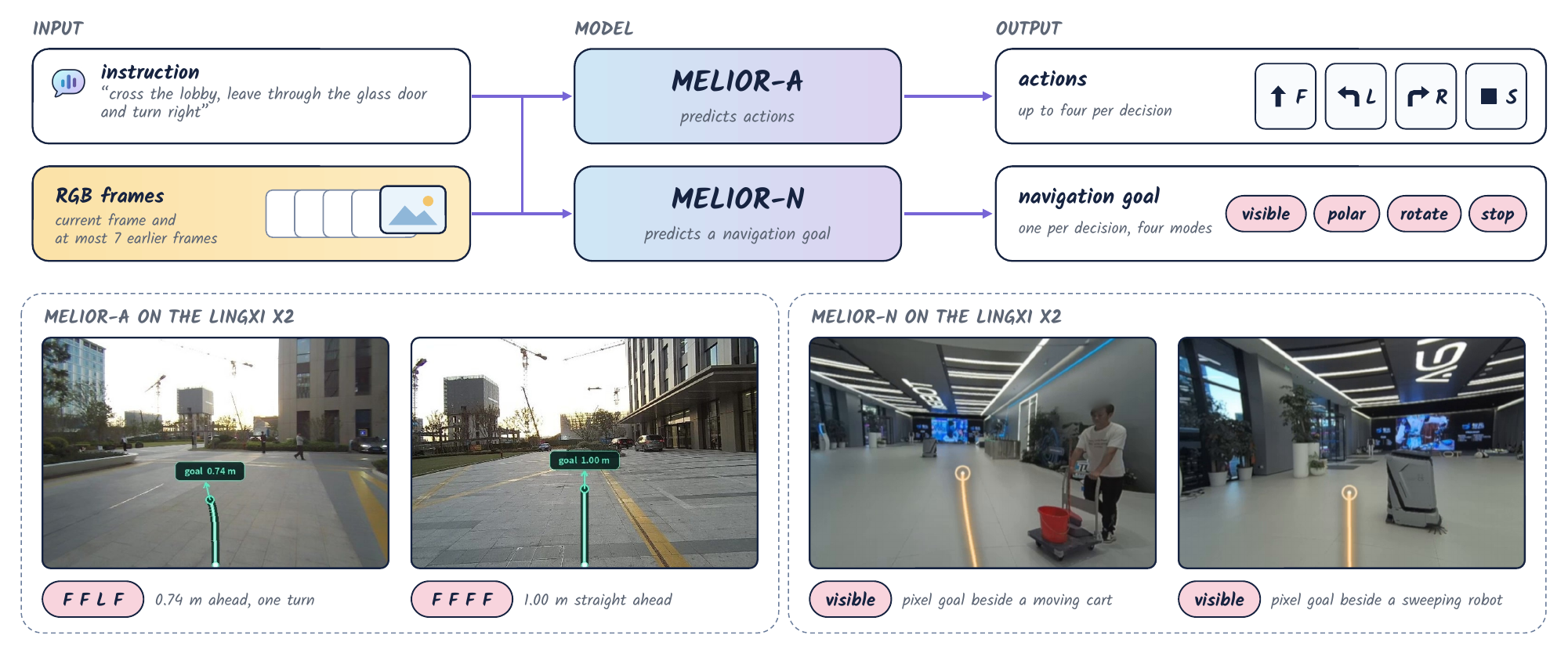}
\caption{\textbf{The two versions of \MEL.}
Both read the instruction and current and history observations of one forward RGB camera.
At each decision step, \MD{} predicts up to four
primitive actions, and \MP{} predicts a single navigation
goal in one of four modes (\cref{sec:mp}). In the action sequence, \texttt{F} is a 0.25\,m forward step,
\texttt{L} and \texttt{R} are $15^\circ$ left and right turns,
respectively, and \texttt{S} stops the robot.
Bottom: decisions of the two versions on the \Xtwo{}, seen by the forward camera.
For \MD{}, the marker shows where the predicted actions lead.
For \MP{}, the circle marks the pixel goal, and the curve from the lower edge to the goal is drawn for display.}
\label{fig:family}
\end{figure}

Each version is an autoregressive vision--language model that receives the instruction $I$, the current forward RGB frame $o_t$ and a history $H_t$ of at most seven sampled earlier frames:
\begin{equation}
  y_t \sim \pi_\theta(\,\cdot \mid I, H_t, o_t\,), \qquad |H_t| \le 7 .
  \label{eq:policy}
\end{equation}
The instruction is a route description, or a sub-task that the planner of \XAgent (\cref{sec:workflow}) has set.
The two versions differ in the output $y_t$ (\cref{fig:family}).
\MD predicts a short sequence of actions.
\MP predicts one navigation goal: a target point that the model marks in the current frame when the point is in view (\cref{sec:mp} gives the other modes), and to which the path of the robot is planned.
The index $t$ counts decisions: one decision covers up to four primitive actions for \MD and one goal for \MP.

\subsection{\MD\label{sec:md}}

\MD predicts $A_t=(a_{t,1},\dots,a_{t,m_t})$ with $m_t\le 4$ primitive actions from $\{\textsc{forward}, \textsc{left}, \textsc{right}, \textsc{stop}\}$; a forward step is 0.25\,m and a turn is $15^\circ$.
These tokens are the action space of the VLN benchmarks, so the model is scored on R2R-CE and RxR-CE as it is.
In simulation the actions are executed one by one.
On the \Xtwo the actions of one inference are composed into a single relative pose target for the locomotion controller, so one inference covers up to one metre of walking.

\subsection{\texorpdfstring{\MP}{\FamilyName-N}\label{sec:mp}}
\begin{figure}[t]
\centering
\includegraphics[width=\linewidth]{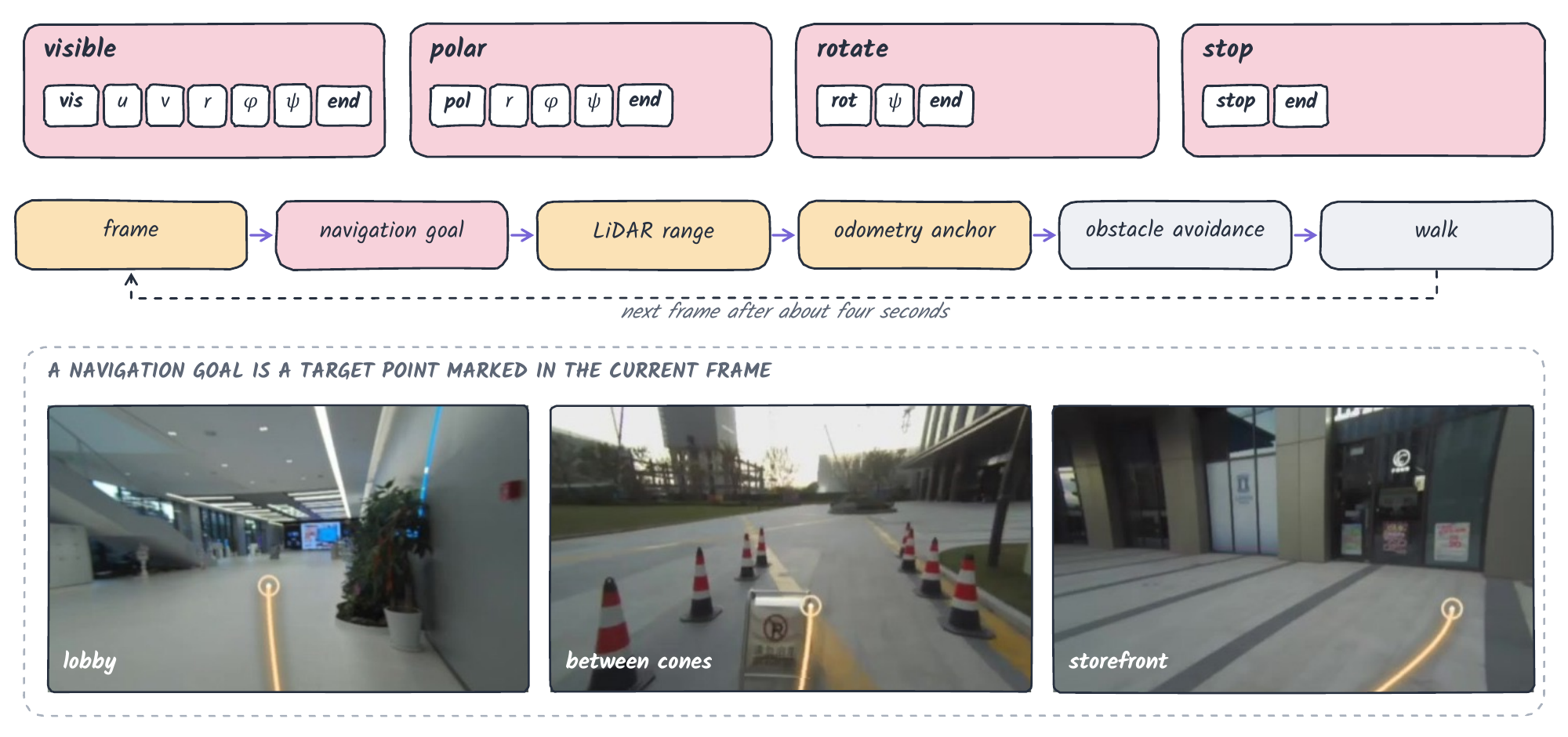}
\caption{\textbf{The navigation-goal interface of \MP.}
Top: the four modes of the navigation command and their tokens.
\textsc{visible} marks a pixel $(u,v)$ of the current view with range $r$, bearing $\phi$ and final heading $\psi$; \textsc{polar} gives a goal outside the view; \textsc{rotate} turns in place; \textsc{stop} declares the instruction complete.
Middle: from one frame to a walking robot.
The model predicts a navigation goal, the LiDAR ranges its pixel, the goal is anchored in odometry, the obstacle-avoidance layer takes the robot around static and dynamic obstacles, and the controller walks; the next frame is taken after about four seconds (\cref{sec:goal_adapter,sec:avoidance}).
Bottom: pixel goals (circles) of \MP on the forward camera of the campus route; the curve from the lower edge to the goal is drawn for display.}
\label{fig:pixelgoal}
\end{figure}

At each decision step, \MP predicts one navigation goal and outputs it as a
navigation command $y_t$, which the
executor translates into low-level actions (\cref{fig:pixelgoal}).
The command takes one of four modes.
\begin{equation}
  \begin{aligned}
    y_t \in \bigl\{&
      \text{\textsc{visible}}(u,v,r,\phi,\psi),\
      \text{\textsc{polar}}(r,\phi,\psi),\ \text{\textsc{rotate}}(\psi),\
      \text{\textsc{stop}}\bigr\}.
  \end{aligned}
  \label{eq:goalblock}
\end{equation}
Each mode is selected by its mode token, \texttt{<visible>}, \texttt{<polar>}, \texttt{<rotate>}, or \texttt{<stop>}.
When the model can point to where the robot should go next in the current image, it outputs \texttt{<visible>} and gives the image coordinates $(u,v)$ as the \emph{pixel goal}
\citep{internrobotics2025internvlan1,wei2026dualvln,majumdar2026robostral}.
When the robot needs to move towards a location that is out of sight, the model outputs \texttt{<polar>}.
In both modes, $r$ and $\phi$ give the planar distance and bearing to
that location, and $\psi$ gives the desired heading on arrival. All
three are relative to the robot's current pose.
When the robot only needs to turn in place, the model outputs
\texttt{<rotate>} with turn angle $\psi$. It outputs \texttt{<stop>} when
it judges that the instruction is complete.
The model emits a mode token followed by any quantised
numeric tokens and an end token.

\paragraph{Execution}
In simulation, the executor back-projects the pixel from a
\texttt{<visible>} command using simulator depth and plans a path on the navigation mesh.
For \texttt{<polar>}, it plans to the position given by $(r,\phi)$ and uses $\psi$ as the desired arrival heading.
For \texttt{<rotate>}, it converts $\psi$ into in-place turns.
On the robot, the goal adapter in \cref{sec:goal_adapter} receives the decoded commands. Further details of the token format, quantisation, and execution in simulation are given in \cref{app:pixelgoal}.

\subsection{Training\label{sec:training}}

We train both versions in two stages; \cref{sec:evaluation} gives the backbones and the training compute.
In the first stage, we perform supervised fine-tuning (SFT) on expert navigation trajectories from R2R, RxR, SRDF, and \XSyn. We also include video and spatial question-answering examples from LLaVA-Video and VSI-590K \citep{zhang2025llavavideo,yang2025cambrians}. \Cref{sec:training_data} gives the composition of the SFT data and the sample counts for \MP{}. In the second stage, we run each model in simulation and use DAgger \citep{ross2011dagger} to collect expert corrections for the states it visits. We then fine-tune the models on these examples to help them recover when they deviate from the demonstrated route.

%% file: sections/05_explicit.tex
\section{\MX\label{sec:explicit}}

\begin{figure}[t]
\centering
\includegraphics[width=\linewidth]{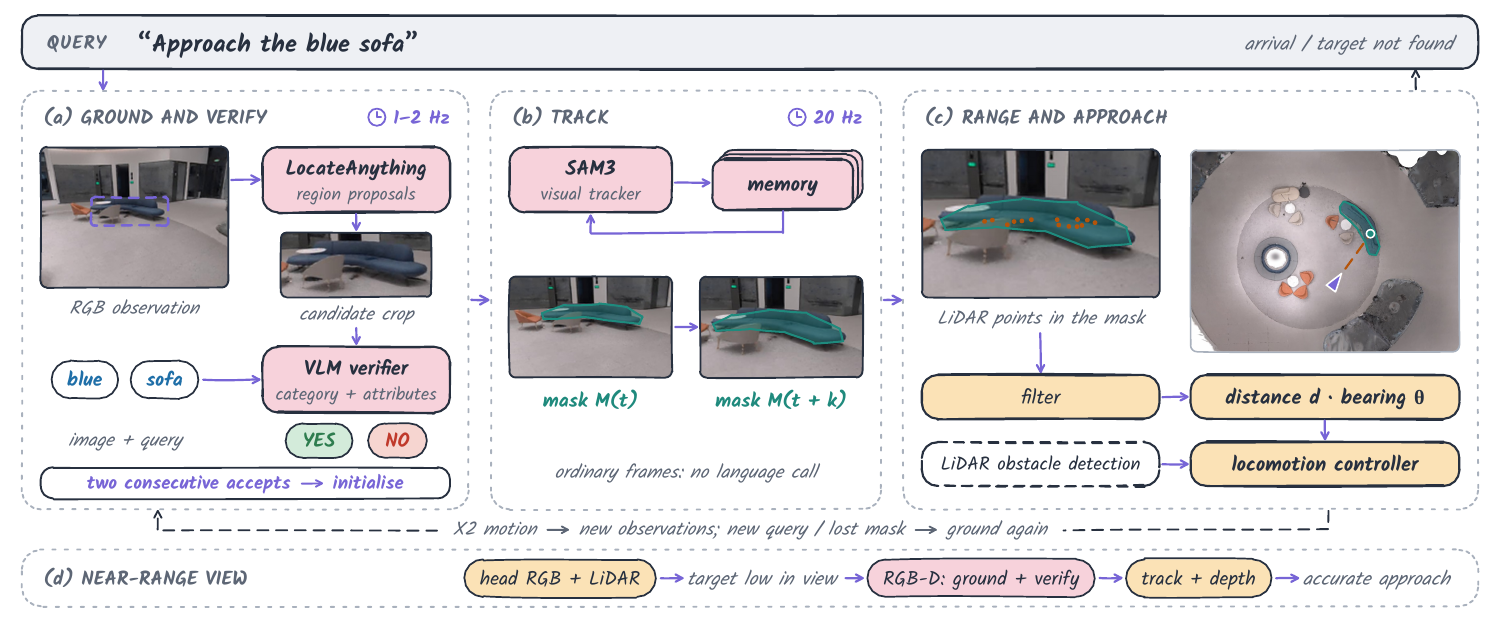}
\caption{\textbf{Overview of \MX.}
Given a language query, the skill (a) grounds and verifies the target, (b) tracks its mask, and (c) estimates its distance $d$ and bearing $\theta$ from the LiDAR points inside the mask to approach it; (d) near the target it switches to the downward RGB-D view.
The tag of a stage gives its update rate.
The images are renderings of the lobby twin with illustrative overlays.}
\label{fig:explicit}
\end{figure}

The last part of a mission is the approach to a named object, such as a vending machine, a sofa or a store entrance.
Object-directed navigation connects an object that is selected in language to the spatial feedback that motion requires.
\MX realises this connection through explicit open-vocabulary semantic tracking.
It is a local skill that acquires a target by its semantic attributes, carries its mask across observations, and converts the tracked region into geometric feedback for the approach (\cref{fig:explicit}).
Given a query $q$, the skill maintains a target mask $M_t$ and an estimated position $\mathbf{p}^B_t$ in the robot base frame.
Its design separates semantic confirmation, visual propagation and geometric control.
Language reasoning selects the target at acquisition and reacquisition; visual memory updates its spatial support on ordinary frames; range measurements convert that support into distance and bearing.
The explicit target state $(M_t,\mathbf{p}^B_t)$ links the queried object to the motion that it guides, and both elements can be inspected afterwards.

\paragraph{Attribute-aware target acquisition}
LocateAnything-3B \citep{wang2026locateanything} predicts up to three candidate boxes from the image and the normalised query.
A compact vision--language verifier (\cref{app:runtime}) then evaluates each candidate crop, together with its local context, against the requested category, colour, appearance, structure and pose.
The locator searches regions with an open vocabulary, and the verifier selects among them by attributes.
For a candidate $b$, the semantic acceptance score is computed from the logits of the first generated token,
\begin{equation}
  s(b,q)=\sigma\big(\ell_{\mathrm{YES}}-\ell_{\mathrm{NO}}\big),
  \label{eq:semantic_score}
\end{equation}
where $\ell_{\mathrm{YES}}$ and $\ell_{\mathrm{NO}}$ are the verifier logits of the two answers given the candidate crop and the query, and $\sigma(z)=(1+e^{-z})^{-1}$.
Candidates with $s\ge 0.5$ are retained and the one with the highest score is chosen; proposals that cover more than 60\% of the image are rejected.

Initialisation and ordinary reacquisition use two consecutive semantic acceptances for the same query.
This check is applied at each observation and places no constraint on the position of the box, which accommodates the image displacement while the robot turns.
The second accepted box initialises the tracker.

\paragraph{Visual memory between grounding events}
The accepted box is the visual prompt for SAM3 \citep{carion2025sam3}, which initialises the target mask and the tracking state.
Its video memory carries the accepted target across changes in appearance and position.
The locator and the verifier run at acquisition events, and ordinary frames use SAM3 propagation alone.
Language reasoning is therefore used only for semantic decisions, while every frame supplies an updated mask for geometric feedback.
A propagated mask carries the latest accepted semantic score, and reacquisition refreshes the semantic validation.
A changed query, an explicit reset, or three consecutive frames without a valid propagated mask starts a new grounding cycle.

\paragraph{Mask-conditioned geometry}
The tracked mask selects the range measurements that connect the semantic target to the controller.
In the standard RGB view, calibrated LiDAR points are projected into the image,
\begin{equation}
  \mathbf{p}_i^C=R_{CL}\,\mathbf{p}_i^L+\mathbf{t}_{CL},\qquad
  \mathbf{u}_i=\pi_{K,D}(\mathbf{p}_i^C),\qquad
  \mathcal{P}_t=\big\{\,i\in\mathcal{V}_t \;\big|\; M_t(\mathbf{u}_i)=1,\ \mathbf{u}_i\in\widetilde{b}_t\,\big\},
  \label{eq:mask_association}
\end{equation}
where $t$ indexes observations, $i$ indexes LiDAR points, and the superscripts $L$, $C$ and $B$ denote LiDAR, camera and robot base coordinates.
The rotation $R_{CL}$ and the translation $\mathbf{t}_{CL}$ map LiDAR points to the camera frame, and $\pi_{K,D}$ projects them to pixels with the intrinsics $K$ and the distortion parameters $D$.
The set $\mathcal{V}_t$ contains the valid points that project into the image, $M_t$ is the binary target mask, and the centrally contracted bounding box $\widetilde{b}_t$ concentrates the association on the interior of the mask.

For the camera-axis depths $z_i^C$, a percentile-based depth window selects a subset $\mathcal{S}_t\subseteq\mathcal{P}_t$ around the foreground depth; when too few points remain, the estimator uses the original candidate set.
At least two points support a position estimate,
\begin{equation}
  \mathbf{p}_t^B=\operatorname*{med}_{i\in\mathcal{S}_t}\big(R_{BL}\,\mathbf{p}_i^L+\mathbf{t}_{BL}\big),
  \label{eq:target_position}
\end{equation}
where $R_{BL}$ and $\mathbf{t}_{BL}$ map LiDAR points to the base frame and $\operatorname{med}$ is the coordinate-wise median.
Mask support, interior sampling and depth selection focus the estimate on the target region, and the base-frame position gives the distance and bearing used for the approach.

\paragraph{Complementary views near the target}
On the \Xtwo, the head-mounted camera provides the standard RGB view, and a downward-pitched RGB-D camera observes low objects near the feet of the robot.
The skill switches the view when the target approaches the lower boundary of the head-camera image.
The projected prior position of the target forms the request for the view switch.
In the first RGB-D view the target is grounded and verified afresh, and SAM3 then propagates its mask.
Aligned depth samples inside the mask are back-projected and transformed into the base frame for the position estimate.
Only one camera holds the tracking state at a time, and the procedure returns to reacquisition in the RGB view when acquisition or tracking in the RGB-D view fails.

\paragraph{Approach control}
The controller aligns the robot with the bearing of the target and regulates forward motion towards a stand-off distance that depends on the view.
Write $\mathbf{p}_t^B=(x_t,y_t,z_t)$ with $x$ forward and $y$ to the left.
The planar distance and the bearing are $d_t=\sqrt{x_t^2+y_t^2}$ and $\theta_t=\operatorname{atan2}(y_t,x_t)$.
With the distance error $e_t=d_t-d^\star$ and the dead-banded bearing $\bar\theta_t$, which is zero for $|\theta_t|<\delta_\theta$ and equal to $\theta_t$ otherwise, the velocity request is
\begin{equation}
\begin{aligned}
  a_t&=\max\!\big(0.2,\; 1-|\bar\theta_t|/\theta_f\big),\\
  v_t&=\begin{cases}0, & |e_t|\le\delta_d \ \text{or}\ |\bar\theta_t|\ge\theta_f,\\[2pt]
       \operatorname{clip}(k_v\,e_t\,a_t,\,0,\,v_{\max}), & \text{otherwise},\end{cases}\\
  \omega_t&=\operatorname{clip}(k_\omega\,\bar\theta_t,\,-\omega_{\max},\,\omega_{\max}),
\end{aligned}
  \label{eq:approach_control}
\end{equation}
where $v_t$ and $\omega_t$ are the forward and yaw velocities, $k_v$ and $k_\omega$ are proportional gains, $v_{\max}$ and $\omega_{\max}$ are speed limits, $\delta_d$ and $\delta_\theta$ are the dead bands of distance and bearing, $\theta_f$ is the half-angle of the forward cone, and $\operatorname{clip}(z,l,u)=\min(u,\max(l,z))$.
The factor $a_t$ slows the forward motion as the bearing error grows.
The controller turns in place for large bearing errors and sets the forward request to zero at the stand-off distance; LiDAR obstacle detection can additionally stop the forward motion.
The motion of the robot produces new observations, which closes the loop through mask propagation or reacquisition.

\paragraph{A callable skill}
\MX is called by the planner of \XAgent (\cref{sec:workflow}).
The caller supplies the query and receives an arrival or a target-not-found status, and the skill maintains the target state $(M_t,\mathbf{p}_t^B)$ that is used for motion.
In summary, the target is selected by semantic verification, maintained across observations by visual memory, and approached with geometric feedback.

%% file: sections/06_workflow.tex
\section{\XAgent: The Agentic Workflow\label{sec:workflow}}

\begin{figure}[!ht]
\centering
\includegraphics[width=\linewidth]{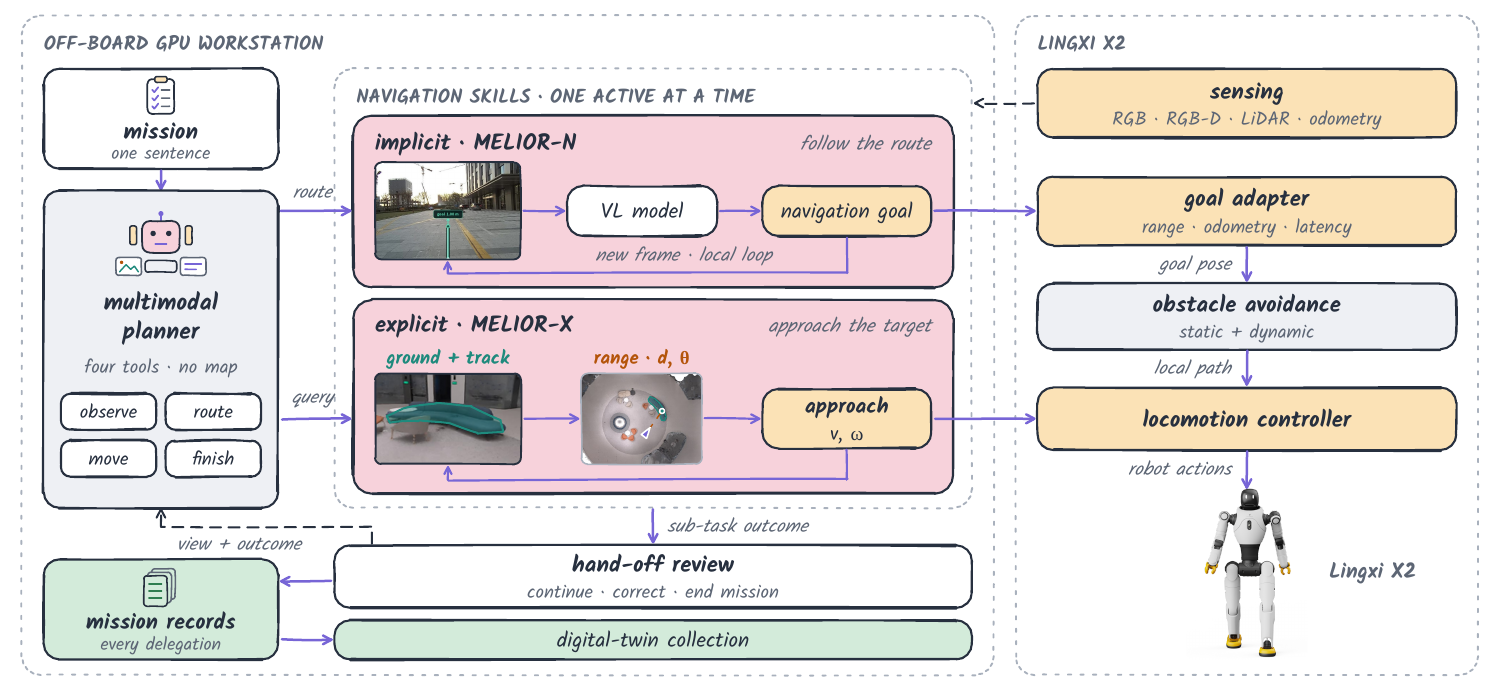}
\caption{\textbf{\XAgent, the agentic workflow of \XNav.}
The planner divides the mission instruction into sub-tasks and reviews their outcome at hand-off.
One navigation skill is active at a time: \MP follows a route, and \MX approaches a named target.
On the robot, the goal adapter converts a navigation goal into a goal pose, and the obstacle-avoidance layer passes it to the locomotion controller.}
\label{fig:runtime}
\end{figure}

An instruction from a person is usually brief and underspecified.
It names a destination or a purpose, for example buying a bottle of water at the vending machine, and leaves the route, the intermediate steps and the stopping place open.
\XAgent is the agentic workflow of \XNav that turns such an instruction into autonomous navigation on the \Xtwo (\cref{fig:runtime}).
A planner divides the instruction into sub-tasks and delegates each of them to a navigation skill, \MEL for a route (\cref{sec:models}) and \MX for a named target (\cref{sec:explicit}); on the robot, each navigation goal passes through a goal adapter and an obstacle-avoidance layer to the locomotion controller.
\XAgent uses no map of the site and no reference route.
Its planner builds on the publicly released runtime of NavGPT-3 \citep{zhou2026navgpt3}.

\paragraph{Planner and navigation skills}\label{sec:workflow_mission}
The planner is a multimodal language model that receives the mission instruction and forward RGB frames of the robot.
It sets one sub-task at a time: a route sub-task, written as one sentence, is delegated to \MP, and a target query to \MX.
The planner is not called inside the loop of a navigation skill, so the robot moves at the rate of the active skill.
A skill hands control back when its sub-task ends, with a stop predicted by \MP or an arrival reported by \MX, or when it reaches a limit on the number of goals or on time.
At hand-off the planner reviews the outcome on the current frame and continues with the next sub-task, corrects the current one with a rewritten instruction, or ends the mission.
A stop of the model therefore ends a sub-task, and the planner decides when the mission is complete.
A new instruction from the person replaces the current mission at any time.

\paragraph{Goal adapter}\label{sec:goal_adapter}
A navigation goal is predicted on a frame that is already outdated when the prediction arrives, because the robot continues to walk.
For a pixel goal $(u,v)$, the goal adapter reads the LiDAR range along the ray through the pixel, which gives a point $g^{(t_0)}$ in the body frame at the capture time $t_0$.
Let $T_{t}$ denote the odometry pose of the body at time $t$.
The goal is fixed in the odometry frame at the capture time and re-expressed in the body frame when it is published at $t_1$:
\begin{equation}
  g^{(t_1)} = T_{t_1}^{-1}\, T_{t_0}\, g^{(t_0)} .
  \label{eq:latency}
\end{equation}
The inference time and the network delay therefore do not displace the goal while the robot moves.
The adapter publishes one goal pose at a time, and a newer goal replaces the previous one (\cref{fig:pixelgoal}, middle row); \cref{app:runtime} lists the settings.

\paragraph{Obstacle avoidance}\label{sec:avoidance}
The route of a mission is shared with pedestrians, carts, service robots, signs and traffic cones.
\XAgent handles such obstacles at three levels, none of which uses a map of the site.
The planner sees the forward frame at hand-off and can name an obstacle and the side on which to pass in the next sub-task.
\MP is trained on reference routes that are planned with a clearance preference (\cref{sec:collection}), and its pixel goals lie on free ground (\cref{fig:pixelgoal}, bottom).
Between the goal pose and the locomotion controller, an obstacle-avoidance layer uses the range sensing of the robot: it plans a local path around static obstacles, yields to moving obstacles, and stops the robot when the path is blocked.
The approach of \MX uses its own LiDAR obstacle detection (\cref{sec:explicit}).

%% file: sections/07_evaluation.tex
\section{Experiments\label{sec:evaluation}}

We evaluate \XNav on the robot and in simulation.
\Cref{sec:eval_mission} reports long-horizon missions on the \Xtwo that lead from indoors to outdoors, with both versions of \MEL.
\Cref{sec:eval_r2s2r} evaluates site adaptation with digital-twin collection, from the scan of an office lobby to the robot in that lobby.
\Cref{sec:eval_benchmark} reports both versions of \MEL on the R2R-CE and RxR-CE benchmarks.

\paragraph{Implementation details}
\MD is built on Qwen3-VL-8B \citep{bai2025qwen3vl} and \MP on Qwen3.5-4B \citep{qwen2026qwen35}; the verifier of \MX is Qwen3-VL-2B.
Both versions of \MEL are trained in two stages (\cref{sec:training}) on NVIDIA H20 GPUs, with about 4,450 GPU hours for \MD and 3,115 GPU hours for \MP.
On the robot, inference runs on an off-board GPU (\cref{sec:platform}).

\subsection{Long-horizon missions on the \Xtwo\label{sec:eval_mission}}

The mission leads from the lobby of an office building, through a door and along a road, to a convenience store on the ground floor of a neighbouring building, about a hundred metres in total.
The robot receives an instruction and the RGB frames of its built-in camera; it has no map of the site and no reference route.
We report two runs on this route.
In the first, \MD follows the instruction directly.
In the second, \XAgent runs \MP on the route with static and dynamic obstacles.

\begin{figure}[t]
\centering
\includegraphics[width=\linewidth]{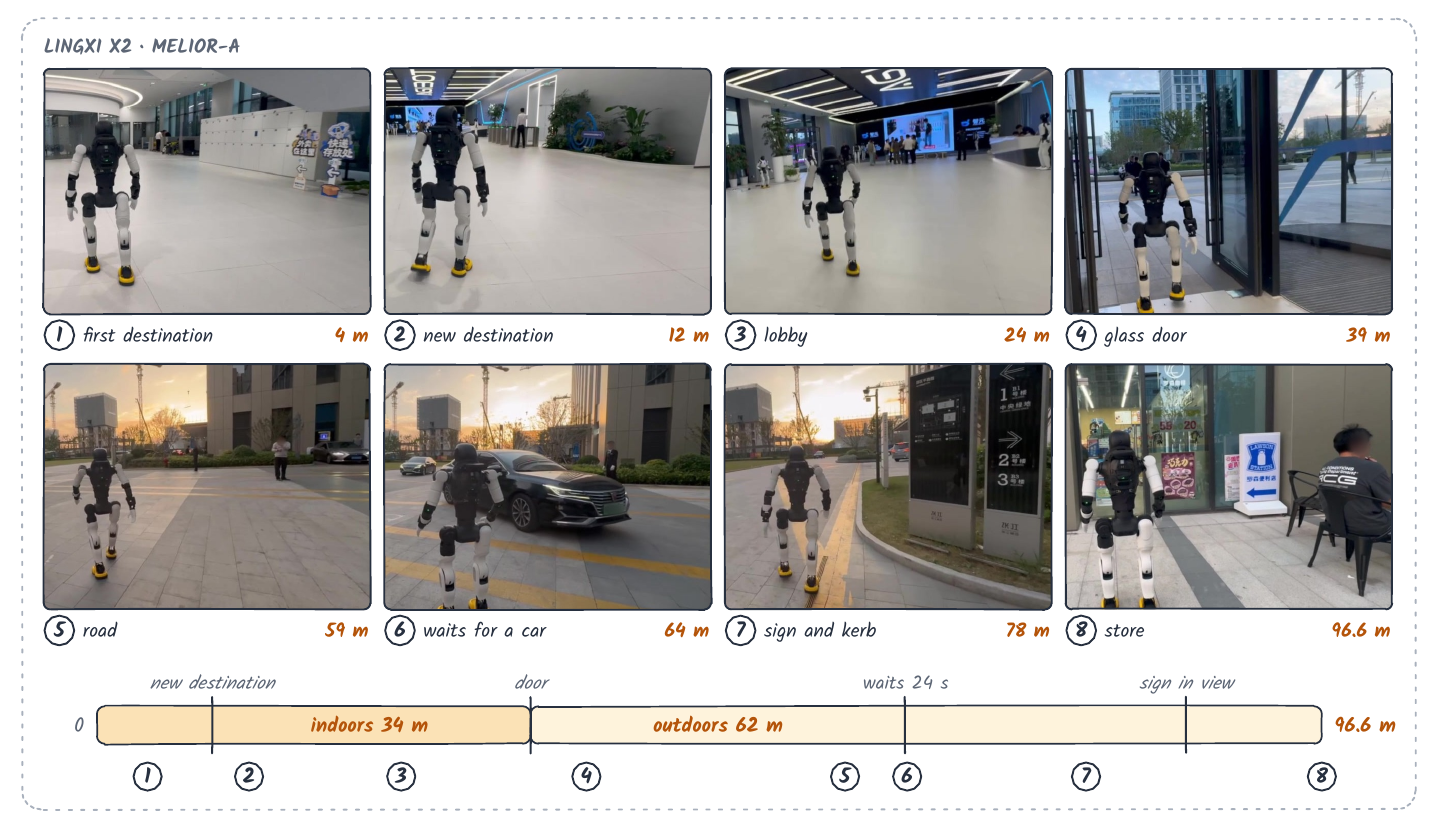}
\caption{\textbf{The mission with \MD.}
The \Xtwo starts in an office lobby (1) and receives a new destination after 9.1\,m, a convenience store in a neighbouring building (2).
It crosses the lobby (3), leaves through a glass door (4), turns right along the road (5), waits while a car and pedestrians cross (6), and stops at the store entrance after 96.6\,m (8).
The bar gives the positions of the frames on the route; distances are cumulative odometry.}
\label{fig:mission}
\end{figure}

\begin{conditionscard}[Setting of the mission with \MD]
\textbf{Robot}\; \Xtwo humanoid.\quad
\textbf{Model}\; \MD; the actions of one decision are executed as one relative pose target.\quad
\textbf{Input}\; the instruction and RGB frames from the built-in camera.\quad
\textbf{Site prior}\; none: no map and no reference route.\quad
\textbf{Operator}\; gives the instruction and replaces it once\MissionTakeovers.
\end{conditionscard}

\paragraph{Mission with \MD}
\Cref{fig:mission} shows the run.
The robot covers 96.6\,m in 3\,min 24\,s, 34.2\,m indoors and 62.4\,m outdoors, several times the ten to fifteen metres of a benchmark episode.
The run contains four events.

\emph{Change of destination.}
The robot has walked 9.1\,m towards a vending machine when the operator sends a new instruction with a destination in another building.
The instruction describes the way in words: across the lobby, through the glass door, to the right, to the store at the foot of the building ahead.
It is the only description of the route that the model receives, and the model turns around and follows it.

\emph{Transition from indoors to outdoors.}
At 34.2\,m the robot leaves the lobby through a glass door.
The view changes from an interior with walls a few metres away to low evening sun, open ground and buildings tens of metres away, and the same model continues outdoors.

\emph{Traffic.}
Between 63.7 and 65.5\,m the robot waits for 24\,s while a car and pedestrians cross, and then continues.

\emph{Arrival.}
The storefront sign comes into view at 85.9\,m, and the robot stops at the entrance at 96.6\,m.

\begin{figure}[t]
\centering
\includegraphics[width=\linewidth]{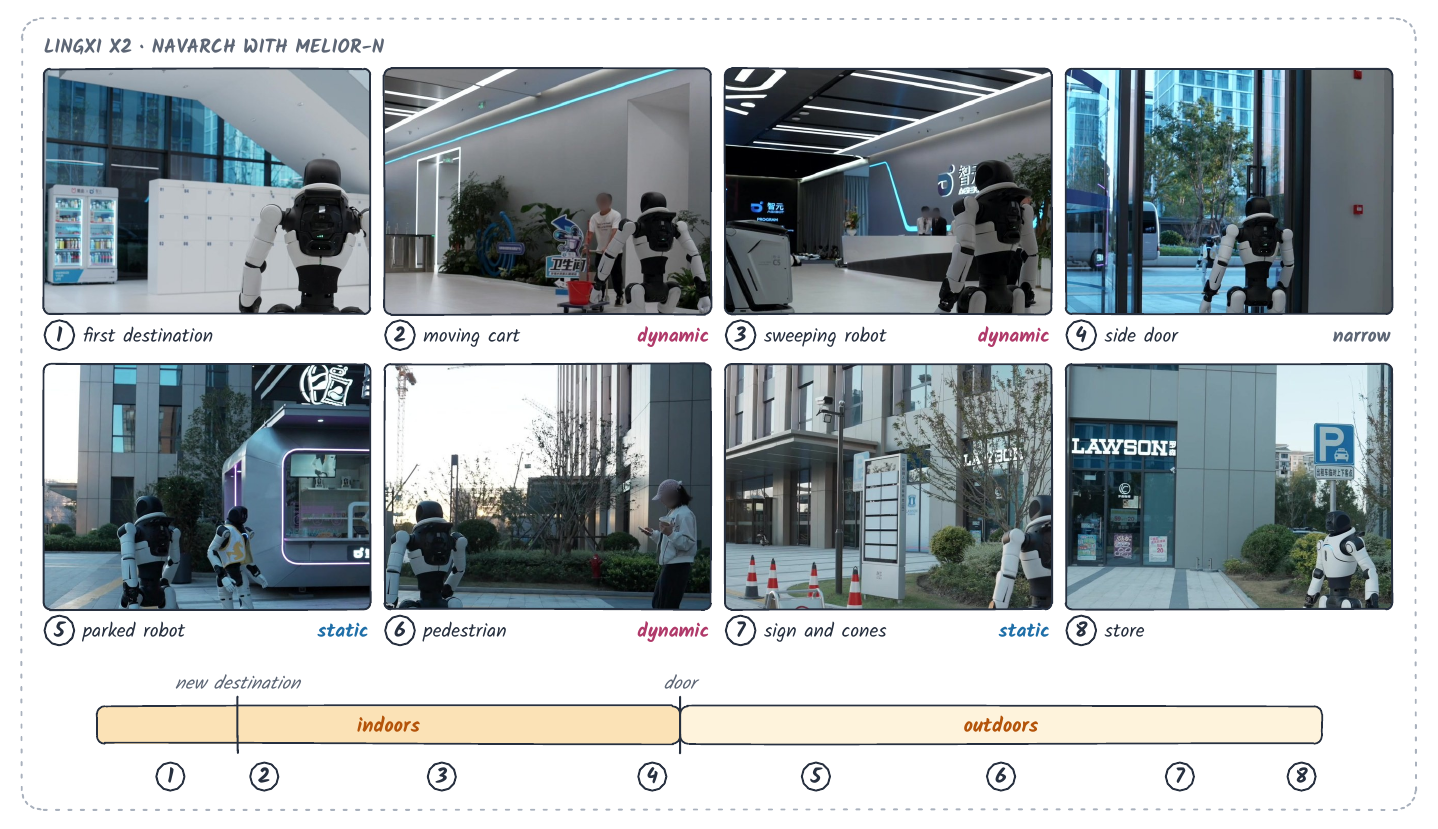}
\caption{\textbf{The mission with \XAgent and \MP, with obstacles on the route.}
\XAgent receives the mission instruction, replaces the destination when a new instruction arrives (between 1 and 2), and drives the \Xtwo with the navigation goals of \MP.
The robot passes a cart that is pushed across the lobby (2) and a sweeping robot (3), walks through the side door (4), passes a robot parked at a kiosk (5), yields to a pedestrian (6), and passes through the gap between a sign and traffic cones (7) before it reaches the store (8).
The note on the right of a frame gives the type of obstacle.}
\label{fig:agentrun}
\end{figure}

\paragraph{Mission with \XAgent and \MP}
In the second run, \XAgent takes the robot from the mission instruction to the destination (\cref{fig:agentrun}).
The operator gives one instruction, ``Go to the vending machine and buy a bottle of water.''
Shortly after the start a new instruction arrives: ``Go outside and find the Lawson to buy a coffee; it is at the second building after you turn right out of the door.''
Neither instruction describes the route step by step.
The planner sets the sub-tasks, to update the destination, to find the side door and to turn right outside, and \MP answers each sub-task with navigation goals.

\begin{conditionscard}[Setting of the mission with \XAgent and \MP]
\textbf{Robot}\; \Xtwo humanoid.\quad
\textbf{System}\; \XAgent with \MP: planner, one navigation goal per decision, goal adapter and obstacle-avoidance layer.\quad
\textbf{Input}\; the mission instruction and RGB frames from the built-in camera; LiDAR and odometry for the execution of goals.\quad
\textbf{Site prior}\; none: no map and no reference route.\quad
\textbf{Operator}\; gives the mission instruction and replaces it once.
\end{conditionscard}

The route contains three dynamic obstacles, two static obstacles and two narrow passages (\cref{fig:agentrun}).
The three levels of obstacle handling (\cref{sec:avoidance}) act together.
The planner names the obstacle and the passing side in the next sub-task, \MP places its goal on free ground, for example in the gap between the cones (\cref{fig:pixelgoal}, bottom), and the obstacle-avoidance layer guides the robot around the cart and the sweeping robot and lets the pedestrian pass.
The robot reaches the store without a map of the site and without a reference route. Together, the two runs show that both versions of \MEL take the \Xtwo through a mission of about a hundred metres across the indoor--outdoor boundary, and that \XAgent completes the mission from a brief instruction in the presence of static and dynamic obstacles.

\subsection{Site adaptation with digital-twin collection\label{sec:eval_r2s2r}}

Digital-twin collection adapts the model to a deployment site: the site is scanned, the model is adapted with data collected in the twin, the adapted model is evaluated in closed loop in the twin, and it is then deployed on the \Xtwo at the same site.
We run this procedure with \MD in the lobby of an office building, the office lobby of \cref{sec:r2s2r}, with the vending machines of the lobby as the target.

\paragraph{Setup}
The twin renders RGB observations with 3D Gaussian splatting, and a co-registered mesh provides depth, collision geometry and occupancy; these are used for collection and are not inputs of the model.
A reference planner generates 500 tasks with collision-free paths of at most 35\,m: 300 instruction-following, 100 target-navigation and 100 demand-navigation tasks.
The tasks are divided into 400 training, 50 validation and 50 test episodes.
\MD is kept frozen, and a low-rank adapter (LoRA, rank 64; \citealp{hu2022lora}) is trained on windows that pair the current frame and at most seven earlier frames with the next four actions.
We compare two adapters.
The first is trained on the 32,404 windows from reference trajectories only.
The second is trained on all 48,006 windows, which add 9,601 recovery windows and 6,001 stop-boundary windows, the correction windows of \cref{sec:r2s2r}.
An episode is a success at radius $r$ if it ends within $r$ of the goal (SR@$r$), and a success with a stop if it also ends with a predicted stop (STOP-SR@$r$); \cref{app:protocol} gives the protocol.

\input{tables/tab_twin}

\paragraph{Results in the twin}
\Cref{tab:twin} reports the results on the 50 validation and 50 test episodes (see also \cref{fig:s2r2s}, panel 4).
With correction windows, SR@3\,m rises from 40\% to 80\% and STOP-SR@3\,m from 34\% to 78\%, and the navigation error falls from 9.63\,m to 3.36\,m.
Episodes that end in a sustained collision fall from 52 to 3 of 100.
Stopping close to the goal pose does not improve: SR@1\,m is 28\% without and 20\% with correction windows.
The correction windows thus reduce collisions and bring the stop into the vicinity of the target, whereas precise stopping at the goal pose remains open.
The scene and the target are the same in training and evaluation, so the result measures adaptation to this site.

\begin{figure}[!ht]
\centering
\includegraphics[width=\linewidth]{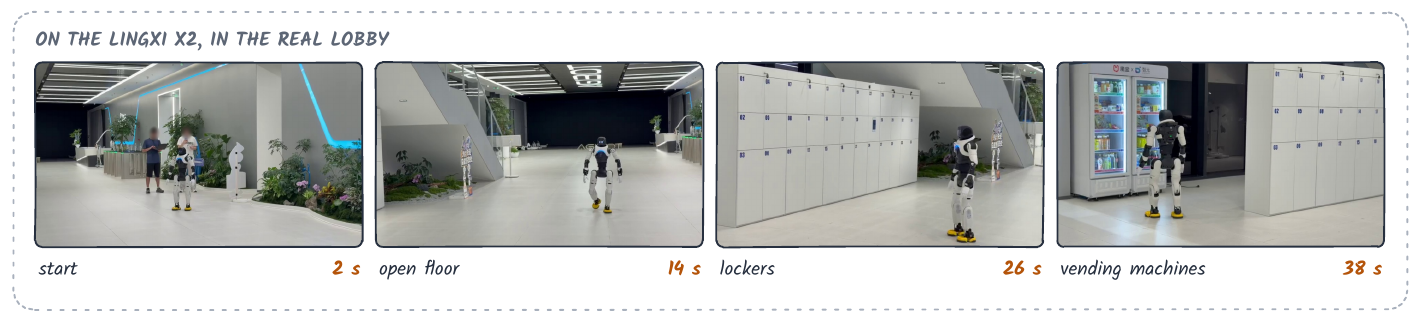}
\caption{\textbf{The \Xtwo with the adapted model in the real lobby.}
The robot crosses the open floor, passes the lockers and reaches the vending machines; the number on the right of a frame is the time since the start.}
\label{fig:siteloop}
\end{figure}

\paragraph{On the robot}
The adapted model is deployed on the \Xtwo in the real lobby.
The robot crosses the open floor of the lobby and approaches the vending machines (\cref{fig:siteloop}).

\subsection{Benchmark results\label{sec:eval_benchmark}}

\begin{figure}[t]
\centering
\includegraphics[width=0.47\linewidth]{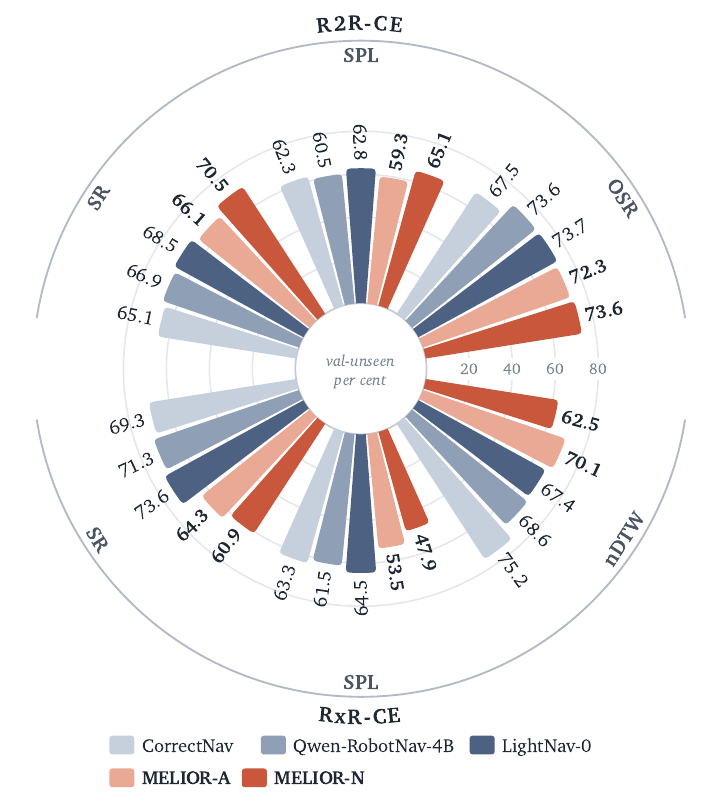}
\caption{\textbf{R2R-CE and RxR-CE val-unseen.}
\MD and \MP compared with three methods of the upper group of \cref{tab:benchmark}; in each group the bars follow the order of the legend.}
\label{fig:glance}
\end{figure}

\Cref{tab:benchmark} reports the val-unseen splits of R2R-CE and RxR-CE, and \cref{fig:glance} compares the two versions of \MEL with three recent methods; \cref{app:protocol} gives the protocol.
On R2R-CE, \MP reaches a success rate of 70.5\%, a navigation error of 3.34\,m and an SPL of 65.1\%, the best values among the methods with one forward RGB stream; its OSR of 73.6\% is 0.1 points below the highest value (LightNav-0, 73.7\%).
\MD reaches 66.1\% on R2R-CE and 64.3\% on RxR-CE.
On RxR-CE, \MP reaches 60.9\%, below the strongest methods of this group.
Methods that use panoramic input, a planner with map tools or training at a much larger scale reach higher success rates and are listed in the lower part of the table for reference.

\input{tables/tab_benchmark}

%% file: tables/tab_twin.tex
\begin{table}[!ht]
\centering\small
\caption{\textbf{Closed-loop results in the digital twin of the lobby (100 episodes).}
SR@$r$: episodes that end within $r$ of the goal; STOP-SR@$r$: episodes that also end with a predicted stop; SPL and STOP-SPL weight the two by path length.
NE in metres, all other columns in per cent.
Both rows are adapters of \MD.}
\label{tab:twin}
\setlength{\tabcolsep}{4.2pt}
\begin{tabular}{@{}l cc cc c c cc@{}}
\toprule
 & \multicolumn{2}{c}{\tablehead{Within 3\,m}} & \multicolumn{2}{c}{\tablehead{Path-weighted, 3\,m}} & & & \multicolumn{2}{c}{\tablehead{Closer radius}}\\
\cmidrule(lr){2-3}\cmidrule(lr){4-5}\cmidrule(l){8-9}
\tablehead{Training windows} & SR & STOP-SR & SPL & STOP-SPL & nDTW & NE & SR@1\,m & SR@0.5\,m\\
\midrule
Reference trajectories only & 40 & 34 & 30.9 & 28.1 & 29.2 & 9.63 & 28 & 13\\
\oursrow With correction windows & 80 & 78 & 79.5 & 77.5 & 66.4 & 3.36 & 20 & 3\\
\bottomrule
\end{tabular}
\end{table}

%% file: tables/tab_benchmark.tex
\begin{table}[!ht]
\centering\small
\caption{\textbf{R2R-CE and RxR-CE, val-unseen.}
NE: navigation error in metres; OSR, SR, SPL and nDTW in per cent.
Results of other methods are taken from the cited papers; a dash marks a value that is not reported.
Bold marks the best value of the upper group; the \MD row is the mean of two checkpoints.
The lower group lists methods with panoramic input, a planner with map tools, or training at a much larger scale.}
\label{tab:benchmark}
\setlength{\tabcolsep}{4.6pt}
\begin{tabular}{@{}l cccc c cccc@{}}
\toprule
 & \multicolumn{4}{c}{\tablehead{R2R-CE}} & & \multicolumn{4}{c}{\tablehead{RxR-CE}}\\
\cmidrule(lr){2-5}\cmidrule(l){7-10}
\tablehead{Method} & NE$\downarrow$ & OSR$\uparrow$ & SR$\uparrow$ & SPL$\uparrow$ & & NE$\downarrow$ & SR$\uparrow$ & SPL$\uparrow$ & nDTW$\uparrow$\\
\midrule
\rowgroup{10}{One forward RGB stream}\\
NaVid \citeyearpar{zhang2024navid} & 5.47 & 49.1 & 37.4 & 35.9 & & -- & -- & -- & --\\
Uni-NaVid \citeyearpar{zhang2025uninavid} & 5.58 & 53.3 & 47.0 & 42.7 & & 6.24 & 48.7 & 40.9 & --\\
NaVILA \citeyearpar{cheng2025navila} & 5.22 & 62.5 & 54.0 & 49.0 & & 6.77 & 49.3 & 44.0 & 58.8\\
StreamVLN \citeyearpar{wei2026streamvln} & 4.98 & 64.2 & 56.9 & 51.9 & & 6.22 & 52.9 & 46.0 & 61.9\\
NavFoM \citeyearpar{zhang2026navfom} & 5.01 & 64.9 & 56.2 & 51.2 & & 5.51 & 57.4 & 49.4 & 60.2\\
JanusVLN \citeyearpar{zeng2026janusvln} & 4.78 & 65.2 & 60.5 & 56.8 & & 6.06 & 56.2 & 47.5 & 62.1\\
DualVLN \citeyearpar{wei2026dualvln} & 4.05 & 70.7 & 64.3 & 58.5 & & 4.58 & 61.4 & 51.8 & 70.0\\
CorrectNav \citeyearpar{yu2026correctnav} & 4.24 & 67.5 & 65.1 & 62.3 & & 4.09 & 69.3 & 63.3 & \textbf{75.2}\\
Qwen-RobotNav-4B \citeyearpar{zhang2026qwenrobotnav} & 4.22 & 73.6 & 66.9 & 60.5 & & 4.15 & 71.3 & 61.5 & 68.6\\
Qwen-RobotNav-8B \citeyearpar{zhang2026qwenrobotnav} & 4.36 & 72.7 & 65.7 & 59.6 & & 4.16 & 73.4 & 63.5 & 69.9\\
LightNav-0 \citeyearpar{lightorigins2026lightnav0} & 3.91 & \textbf{73.7} & 68.5 & 62.8 & & \textbf{3.66} & \textbf{73.6} & \textbf{64.5} & 67.4\\
\oursrow \MD & 3.86 & 72.3 & 66.1 & 59.3 & & 4.33 & 64.3 & 53.5 & 70.1\\
\oursrow \MP & \textbf{3.34} & 73.6 & \textbf{70.5} & \textbf{65.1} & & 4.80 & 60.9 & 47.9 & 62.5\\
\midrule
\rowgroup{10}{Reference: training at a much larger scale, panoramic input, or a planner with map tools}\\
Qwen-RobotNav-8B, panoramic \citeyearpar{zhang2026qwenrobotnav} & 3.53 & 78.5 & 72.1 & 66.6 & & 3.58 & 76.5 & 65.7 & 72.5\\
Robostral Navigate \citeyearpar{majumdar2026robostral} & 3.20 & 81.3 & 77.4 & 74.2 & & 3.47 & 75.1 & 68.7 & --\\
NavGPT VLA-8B \citeyearpar{zhou2026navgpt3} & 2.97 & 82.5 & 76.9 & 70.6 & & 2.65 & 82.4 & 70.6 & 76.7\\
NavGPT-3, planner and VLA \citeyearpar{zhou2026navgpt3} & 2.18 & 87.0 & 81.5 & 70.4 & & 1.56 & 90.4 & 74.3 & 78.5\\
\bottomrule
\end{tabular}
\end{table}

%% file: sections/09_related.tex
\let\XNavFloatBarrier\FloatBarrier
\let\FloatBarrier\relax
\section{Related Work\label{sec:related}}
\let\FloatBarrier\XNavFloatBarrier

\paragraph{Vision-language navigation}
VLN moved from discrete graphs \citep{anderson2018vln,ku2020rxr} to continuous environments \citep{krantz2020vlnce}, and from task-specific models to vision--language models that read the camera stream directly \citep{zhang2024navid,zhang2025uninavid,cheng2025navila,zhang2026navfom}.
Recent models increase the scale of training: Qwen-RobotNav trains one model with a parameterised interface on 15.6M samples \citep{zhang2026qwenrobotnav}, Robostral Navigate trains on 2.4M simulated trajectories and adds reinforcement learning \citep{majumdar2026robostral}, and LightNav-0 builds a compact generalist navigation model \citep{lightorigins2026lightnav0}.
The models differ in their output: primitive actions \citep{zhang2024navid,wei2026streamvln}, mid-level language actions \citep{cheng2025navila}, metric waypoints \citep{zhang2026navfom,zhang2026qwenrobotnav}, waypoints pointed at in the image \citep{majumdar2026robostral}, or a pixel goal that a second, faster model converts into a trajectory \citep{internrobotics2025internvlan1,wei2026dualvln}.
\MEL takes the same type of input, RGB frames from a single forward camera, and provides two outputs under one input contract: \MD predicts primitive actions, the action space of the VLN benchmarks, and \MP predicts a navigation goal.

\paragraph{Navigation on legged and humanoid robots}
NaVILA drives legged robots with mid-level language actions above a locomotion policy and evaluates 25 instructions in workspace, home and outdoor settings, on a quadruped and on a humanoid \citep{cheng2025navila}.
DualVLN runs on wheeled, quadruped and humanoid robots and reports trials in three indoor scenarios \citep{wei2026dualvln}, and TANGO controls a Unitree G1 with a whole-body vision--language--action model along indoor routes of about 10 and 30\,m \citep{chen2026tango}.
Concurrent with this report, LHNav studies humanoid navigation under long-horizon instructions that name several successive destinations and deploys it on a Unitree G1 \citep{wang2026lhnav}, and Compass~1 evaluates one navigation model on wheeled, quadruped, humanoid and aerial robots, from indoor environments to streets with pedestrians \citep{zhang2026compass1}.
\XNav takes a bipedal humanoid through one mission of about a hundred metres that starts indoors and ends outdoors, with static and dynamic obstacles on the route.

\paragraph{Navigation data and sim-to-real transfer}
Training sets have grown with the models, to 15.6M samples for Qwen-RobotNav \citep{zhang2026qwenrobotnav} and 19.3M records for NavGPT-3 \citep{zhou2026navgpt3}.
For correction data, DAgger relabels the states that a policy visits \citep{ross2011dagger}, and CorrectNav builds a self-correction flywheel from the errors of the model on its training trajectories \citep{yu2026correctnav}.
Sim-to-real work for VLN transfers policies through learned 3D representations \citep{wang2025sim2real}, and 3D Gaussian splatting renders a scanned place with the appearance that a robot camera observes \citep{kerbl2023gaussian}.
The data engine of \XNav combines three collection routes: open-source scenes at scale, the digital twin of the deployment site with correction windows, and the real robot.

\paragraph{Open-vocabulary object navigation}
Open-vocabulary grounding and promptable video segmentation make it practical to select an object by a phrase and to track it through a video \citep{liu2024groundingdino,wang2026locateanything,carion2025sam3}.
HoloAgent-0 retrieves a target from a hierarchical scene graph, verifies it online, and explores when the memory does not contain it \citep{zhou2026holoagent}.
\MX works from the live view without a stored map: it separates the semantic decision at acquisition from the mask that visual memory propagates, and it closes the approach loop with range measurements inside the mask.

\paragraph{LLM-based navigation agents}
NavGPT showed that a language model can reason about navigation in text \citep{zhou2024navgpt}; NavGPT-3 connects such a model to an action policy through a runtime with reasoning, acting and monitoring threads \citep{zhou2026navgpt3}, and VbotNav-0 couples a cloud policy with an edge policy along a supplied route \citep{tan2026vbotnav}.
\XAgent builds on the NavGPT-3 runtime and restricts the planner to the input of a single forward camera, without a map.
It adds the components for deployment on a walking humanoid: two navigation skills behind one interface, a goal adapter with latency compensation, and an obstacle-avoidance layer between the goal pose and the locomotion controller.

%% file: sections/10_conclusion.tex
\section{Conclusion\label{sec:conclusion}}

We presented \XNav, a navigation system that takes a bipedal humanoid through a language-instructed long-horizon mission from indoors to outdoors.
Its data engine collects navigation data in open-source scenes, in digital twins of deployment sites and on real robots.
Its navigation model, \MEL, takes only RGB images from the built-in camera of the \Xtwo and predicts actions (\MD) or a navigation goal (\MP).
Its agentic workflow, \XAgent, turns a brief human instruction into sub-tasks, approaches named targets by explicit navigation, and avoids static and dynamic obstacles.
On the \Xtwo, \XNav completes a mission of about a hundred metres from an office lobby to a store in a neighbouring building.
At a deployment site, adaptation with data from the digital twin raises the share of closed-loop episodes that stop within 3\,m of the goal from 34\% to 78\%, and the adapted model is deployed on the robot.

\paragraph{Limitations and future work}
The missions of this report follow one route on one floor; stairs, lifts and doors that have to be opened are outside the present system, and site adaptation is evaluated at one site.
We plan three lines of work.
The first is a data-scaling study that varies data volume, the number of scenes, and the proportions of correction data and real-robot data.
The second is a spatial memory that persists beyond one mission: a scene graph that the robot builds during exploration, keeps between missions and queries by language.
The third is a learned local policy between the navigation goal and the locomotion controller (\cref{app:localpolicy}).

%% file: sections/appendix/B_pixelgoal.tex
\section{Navigation Command Format\label{app:pixelgoal}}

This appendix describes the token format of the navigation commands
predicted by \MP in \cref{sec:mp}.
At each decision step, the model emits one command consisting of
a mode token, the numerical fields required by that mode, and
the end token.

\paragraph{Fields and coordinate conventions}
For \texttt{<visible>}, $(u,v)$ gives the pixel goal in the current
RGB frame. The image origin is at the top-left corner, with $u$
increasing to the right and $v$ increasing downward.
In both \texttt{<visible>} and \texttt{<polar>}, $r$ is the planar
distance to the target, $\phi$ is its bearing, and $\psi$ is the
desired heading on arrival.
These fields are relative to the robot's pose at the current
decision step: $x$ points forward, $y$ points left, and positive
angles are counter-clockwise when viewed from above.
Both $\phi$ and $\psi$ are measured from the current heading.
For example, $\psi=0$ asks the robot to arrive facing its original
direction, even if the target lies to one side.
For \texttt{<rotate>}, $\psi$ specifies the signed angle of an
in-place turn.
The \texttt{<stop>} mode has no numerical fields.

\paragraph{Token vocabulary and encoding}
\Cref{tab:schema} lists the 365 navigation tokens.
Each numerical field is encoded as a single token.
The pixel coordinates $u$ and $v$ share one token set, and the
angles $\phi$ and $\psi$ share another; their positions in the
command determine which field each token represents.
The token-to-ID mapping and quantisation settings are saved
with each checkpoint.

\begin{table}[ht]
\centering\small
\caption{\textbf{Navigation token vocabulary of \MP.}}
\label{tab:schema}
\begin{tabularx}{\linewidth}{@{}l r >{\raggedright\arraybackslash}X@{}}
\toprule
\tablehead{Group} & \tablehead{Tokens} & \tablehead{Meaning and encoding}\\
\midrule
Control & 5 &
\texttt{<visible>}, \texttt{<polar>}, \texttt{<rotate>},
\texttt{<stop>}, and \texttt{<end>}\\
Pixel & 256 &
Uniformly spaced coordinate values on the $640\times480$ frame,
shared by $u$ and $v$\\
Range & 32 &
Bins uniformly spaced in $\log(1+r)$ over 0--15\,m;
larger distances map to the final bin\\
Angle & 72 &
Circular values at $5^\circ$ intervals, shared by $\phi$ and $\psi$\\
\midrule
Total & 365 & \\
\bottomrule
\end{tabularx}
\end{table}
Pixel coordinates and angles are rounded to the nearest
representable value.
Distances are assigned to a bin in log space and decoded using
that bin's midpoint in log space.
This gives finer distance resolution near the robot and coarser
resolution farther away.

\paragraph{Command format and decoding}
The numerical fields follow a fixed order:
\texttt{<visible>} is followed by $u,v,r,\phi,\psi$;
\texttt{<polar>} by $r,\phi,\psi$; and
\texttt{<rotate>} by $\psi$.
Every command ends with \texttt{<end>}, which immediately follows
the mode token for \texttt{<stop>}.
During autoregressive decoding, the first token selects one of
the four modes.
Each subsequent position is restricted to the token set for
the expected field.
For example, after \texttt{<visible>}, the decoder generates two
pixel tokens, one range token, and two angle tokens.
After the final field, only \texttt{<end>} is allowed.

\paragraph{Execution in simulation}
The executor converts a decoded command into at most four primitive
actions before the model receives a new observation.
It uses the same action space as \MD: a 0.25\,m forward step,
$15^\circ$ left and right turns, and stop.
A target may therefore remain unreached when the next decision begins.
For \texttt{<visible>}, the executor back-projects $(u,v)$ using
the simulator depth and camera intrinsics to obtain a 3D target.
It maps this target onto the navigation mesh and uses a
shortest-path follower to move towards it.
This branch determines the target from the pixel and depth;
the predicted $r$, $\phi$, and $\psi$ are not used for execution.
For \texttt{<polar>}, the executor plans to the position given by
$(r,\phi)$ and uses $\psi$ as the desired arrival heading.
Once the position is reached, any remaining action budget can
be used to turn towards that heading.
For \texttt{<rotate>}, the executor converts $\psi$ into in-place
turns.
A \texttt{<stop>} command produces the stop action.
The model receives RGB observations; simulator depth and the
navigation mesh are used by the executor.

%% file: sections/appendix/C_runtime.tex
\section{Navigation Skills and Settings\label{app:runtime}}

\Cref{tab:modes} compares the two types of sub-task of \cref{sec:modes} and the navigation skills that execute them.
\Cref{tab:runtime} lists the settings of \XAgent with \MP on the \Xtwo, and \cref{tab:explicit_settings} the settings of \MX that are specified in \cref{sec:explicit}.

\begin{table}[ht]
\centering\small
\caption{\textbf{The two types of sub-task and the navigation skills that execute them.}}
\label{tab:modes}
\begin{tabularx}{\linewidth}{@{}l >{\raggedright\arraybackslash}X >{\raggedright\arraybackslash}X@{}}
\toprule
 & \tablehead{Following a route (implicit navigation)} & \tablehead{Approaching a target (explicit navigation)}\\
\midrule
Sub-task & a route described in language & a named object to approach\\
Component & \MD, \MP & \MX\\
Target representation & implicit, in the weights and the context of the model & explicit state: mask $M_t$ and base-frame position $\mathbf{p}^B_t$\\
Input & instruction; RGB only: the current frame and at most seven earlier frames & query; RGB with LiDAR, or RGB-D at close range\\
Language model & runs at every decision & runs when the target is acquired or reacquired\\
Output & up to four actions, or one navigation goal & distance and bearing, then a velocity request\\
Obstacles & obstacle-avoidance layer between the goal pose and the locomotion controller & LiDAR obstacle detection\\
Stop & predicted by the model, reviewed by the planner & the stand-off distance is reached\\
Share of a mission & the route, tens of metres & the last metres\\
\bottomrule
\end{tabularx}
\end{table}

\begin{table}[ht]
\centering\small
\caption{\textbf{\XAgent on the \Xtwo: planner and execution of navigation goals.}}
\label{tab:runtime}
\begin{tabularx}{\linewidth}{@{}l >{\raggedright\arraybackslash}X@{}}
\toprule
\tablehead{Item} & \tablehead{Setting}\\
\midrule
Planner tools & \texttt{observe\_forward}, \texttt{navigate\_by\_instruction}, \texttt{navigate\_relative}, \texttt{terminate\_episode}\\
Planner input & mission instruction; one forward RGB frame and an outcome report per tool result\\
Planner model & \PlannerRobot\\
Navigation model & \MP; the current frame and at most seven earlier frames; frame history cleared at each new instruction\\
Hand-off & \texttt{<stop>}; 32 navigation goals; 180\,s; failure; new mission instruction\\
Saved map & not required\\
Goal range & LiDAR range along the ray through the pixel goal, accepted between 0.10 and 8\,m\\
Goal frame & fixed in the odometry frame at capture time, re-expressed at publication (\cref{eq:latency})\\
Turn in place & for \texttt{<rotate>} and \texttt{<polar>} commands, between $15^\circ$ and $90^\circ$\\
Missing range & command discarded, previous goal kept\\
New frame & about every 4\,s\\
Output to the robot & goal poses, through the obstacle-avoidance layer, to the locomotion controller; no velocity commands\\
Command timeout & heartbeat every 0.2\,s; the robot stops 0.8\,s after the last heartbeat\\
Outdated data & observations older than the time limit and requests of cancelled sub-tasks are rejected\\
\bottomrule
\end{tabularx}
\end{table}
\confirm{SC}{confirm the public naming of the planner model (\PlannerRobot) with legal/PR before release}

\begin{table}[h]
\centering\small
\caption{\textbf{\MX: acquisition, tracking and ranging settings.}}
\label{tab:explicit_settings}
\begin{tabularx}{\linewidth}{@{}l >{\raggedright\arraybackslash}X@{}}
\toprule
\tablehead{Item} & \tablehead{Setting}\\
\midrule
Locator & LocateAnything-3B; up to three candidate boxes per image and query\\
Verifier & compact vision--language model (\cref{sec:evaluation}); score from the first-token logits of the two answers (\cref{eq:semantic_score})\\
Acceptance & score of at least 0.5; highest score chosen; boxes that cover more than 60\% of the image rejected\\
Initialisation & two consecutive acceptances for the same query; the second box initialises the tracker\\
Tracker & SAM3 mask propagation on ordinary frames\\
Reacquisition & changed query, explicit reset, or three consecutive frames without a valid mask\\
Range, standard view & LiDAR points inside the mask and the contracted box; depth window; coordinate-wise median of at least two points\\
Range, near view & aligned RGB-D depth inside the mask, back-projected to the base frame\\
Control & turn in place outside the forward cone; proportional approach to a view-dependent stand-off distance; LiDAR obstacle detection\\
Status & arrival, or target not found\\
\bottomrule
\end{tabularx}
\end{table}

%% file: sections/appendix/D_protocol.tex
\section{Evaluation Protocol\label{app:protocol}}

\paragraph{Benchmarks}
R2R-CE and RxR-CE are evaluated on the val-unseen splits in Habitat \citep{savva2019habitat} with Matterport3D scenes \citep{chang2017matterport3d}: 1,839 episodes for R2R-CE, and the 3,669 episodes with English instructions for RxR-CE.
The agent has a forward step of 0.25\,m and turns of $15^\circ$, and an episode succeeds when the agent stops within 3\,m of the destination.
The models receive one forward RGB camera; the panoramic camera is not used.
We report navigation error (NE), success rate (SR), success weighted by path length (SPL), the share of episodes whose trajectory enters the 3\,m success radius (OSR), and normalised dynamic time warping (nDTW; \citealp{ilharco2019ndtw}).

\paragraph{Closed-loop evaluation in the digital twin}
In \cref{sec:eval_r2s2r}, the model receives rendered views of the head camera ($640\times480$), the current frame and at most seven earlier frames.
Up to two of the four predicted actions are executed per decision, and an episode ends with a predicted stop, with a sustained collision or after 500 steps.
SR@$r$ is the share of episodes that end within $r$ of the goal, and STOP-SR@$r$ the share that end within $r$ of the goal with a predicted stop; SPL and STOP-SPL weight the two by path length.
The results are reported on the 50 validation and 50 test episodes of the site.

\paragraph{Missions on the robot}
For the missions of \cref{sec:eval_mission}, distance is the cumulative odometry of the \Xtwo.

%% file: sections/appendix/E_localpolicy.tex
\section{Outlook: A Learned Local Policy\label{app:localpolicy}}

\begin{figure}[!htb]
\centering
\includegraphics[width=\linewidth]{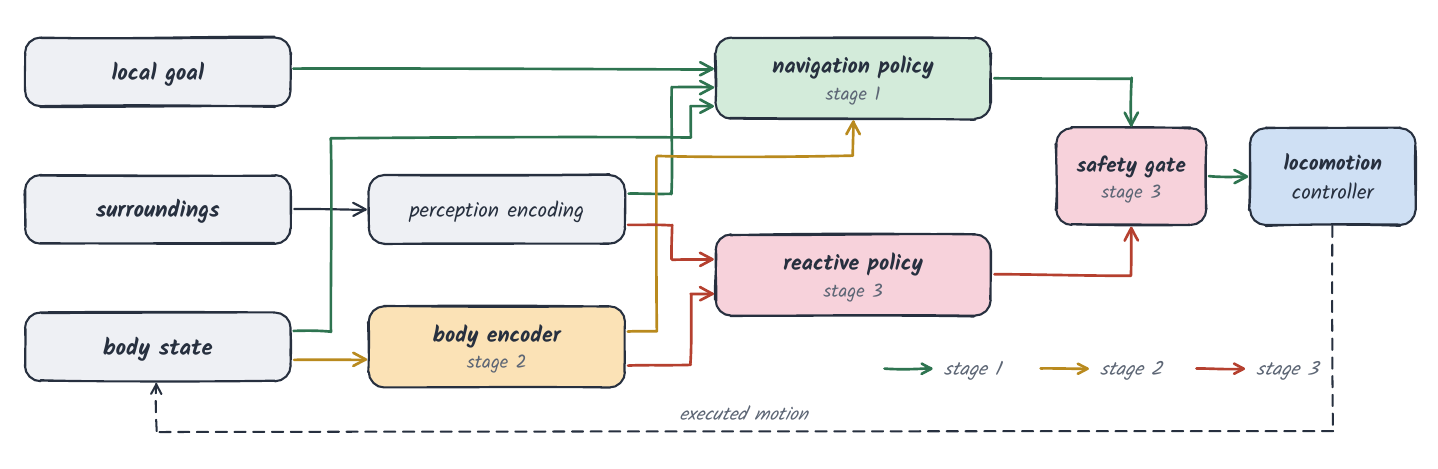}
\caption{\textbf{Design of the learned local policy.}
The policy operates between the navigation goal and the frozen locomotion controller; the colour of an arrow is the stage that adds it.
Stage 1: a navigation policy with one backbone for all robots maps the local goal, the encoded surroundings and the body state to commands.
Stage 2: a body encoder for each type of robot.
Stage 3: a reactive policy with a safety gate for moving obstacles.
The executed motion returns as body state (dashed).}
\label{fig:roadmap}
\end{figure}

At present an obstacle-avoidance layer operates between the navigation goal and the locomotion controller and supports one robot (\cref{sec:avoidance}).
\Cref{fig:roadmap} shows the design of a learned local policy for this position.
It is built in three stages: a navigation policy with one backbone for all robots, which maps the local goal, the encoded surroundings and the body state to commands for a frozen locomotion controller; a body encoder for each type of robot, so that one policy supports more than one robot; and a reactive policy with a safety gate for moving obstacles.

